\documentclass[conference]{IEEEtran}
\usepackage{times}
\let\labelindent\relax

\usepackage[numbers]{natbib}
\usepackage{multicol}
\usepackage{balance}
\usepackage[bookmarks=true]{hyperref}
\usepackage{flushend}

\usepackage{xcolor}

\usepackage{tcolorbox}
\newcommand{\ours}{NaVid}

\newcommand{\egno}{\textit{e}.\textit{g}.} 
\newcommand{\ieno}{\textit{i}.\textit{e}.} 
\newcommand{\etcno}{\textit{etc}} 

\def\CircleArrowright{\ensuremath{%
  \rotatebox[origin=c]{310}{$\circlearrowright$}}}

\newcommand{\vlnbert}{VLN$\protect\CircleArrowright$BERT}

\makeatletter
\def\blfootnote{\xdef\@thefnmark{}\@footnotetext}
\makeatother

\usepackage{mathtools}
\usepackage{amsmath}

\usepackage{amssymb}
\usepackage{amsfonts}
\usepackage{amsopn}
\usepackage{graphicx} 
\usepackage{textcomp}
\usepackage{xfrac}
\usepackage{bbm}
\usepackage{overpic}
\usepackage{subfig}
\usepackage{wrapfig}
\usepackage[ruled,vlined,linesnumbered]{algorithm2e}
\SetAlFnt{\small}
\SetAlCapFnt{\small}
\SetAlCapNameFnt{\small}
\SetAlCapHSkip{0pt}

\usepackage{multirow}
\usepackage{multicol}

\usepackage{booktabs}
\usepackage{footnote}
\usepackage{paralist}
\usepackage{enumitem}
\usepackage{autobreak}
\usepackage{hyperref}
\usepackage{cleveref}

\usepackage{color}
\definecolor{green}{rgb}{0, 0.4, 0}
\definecolor{orange}{rgb}{0.8, 0.6, 0.2}
\definecolor{red}{rgb}{1.0, 0.0, 0.0}
\definecolor{teal}{rgb}{0.0, 0.4, 0.4}
\definecolor{purple}{rgb}{0.65,0,0.65}
\definecolor{saffron}{rgb}{0.95,0.75,0.2}
\definecolor{turquoise}{rgb}{0.0,0.5,0.5}
\definecolor{brown}{rgb}{0.5, 0.16, 0.16}

\usepackage{overpic}
\usepackage{currfile} 

\newlength\savedwidth

\definecolor{lightgray}{rgb}{0.6, 0.6, 0.6}

\newcommand{\jiazhao}[1]{{\textcolor{black}{#1}}}

\usepackage[normalem]{ulem}

\newcommand{\hidecomment}[1]{}

\newcommand{\cO}{\mathcal{O}}
\newcommand{\cA}{\mathcal{A}}

\newcommand{\cI}{\mathcal{I}}

\usepackage{xspace}

\begin{document}

\title{NaVid: Video-based VLM Plans the Next Step\\ for Vision-and-Language Navigation} 

\author{Jiazhao Zhang$^{1,2,\ast}$ \quad\quad Kunyu Wang$^{2,\ast}$ \quad\quad Rongtao Xu$^{2,3,\ast}$ \quad\quad Gengze Zhou$^{4}$ \quad\quad Yicong Hong$^{5}$ \quad\quad \\ Xiaomeng Fang$^{2}$  \quad\quad Qi Wu$^{4}$ \quad\quad Zhizheng Zhang$^{6,\dagger}$ \quad\quad He Wang$^{1,2,6,\dagger}$ \\ \normalsize{ $^1$CFCS, School of Computer Science, Peking University \quad $^2$Beijing Academy of Artificial Intelligence} \\  \normalsize{\quad $^3$CASIA   \quad $^4$University of Adelaide \quad $^5$Australian National University \quad $^6$Galbot} \\ \href{https://pku-epic.github.io/NaVid/}{\texttt{https://pku-epic.github.io/NaVid/} }}

\maketitle
\blfootnote{* Joint first authors (e-mail: zhngjizh@gmail.com). $\dagger$ Corresponding authors (e-mail: zhangzz@galbot.com, hewang@pku.edu.cn).}

\begin{abstract}

Vision-and-language navigation (VLN) stands as a key research problem of Embodied AI, aiming at enabling agents to navigate in unseen environments following linguistic instructions. In this field, generalization is a long-standing challenge, either to out-of-distribution scenes or from Sim to Real. In this paper, we propose NaVid, a video-based large vision language model (VLM), to mitigate such a generalization gap. 
NaVid makes the first endeavor to showcase the capability of VLMs to achieve state-of-the-art level navigation performance without any maps, odometers, or depth inputs. Following human instruction, NaVid only requires an on-the-fly video stream from a monocular RGB camera equipped on the robot to output the next-step action. 
Our formulation mimics how humans navigate  
and naturally gets rid of the problems introduced by odometer noises, and the Sim2Real gaps from map or depth inputs. Moreover, our video-based approach can effectively encode the historical observations of robots as spatio-temporal contexts for decision making and instruction following. We train NaVid with 510k navigation samples collected from continuous environments, including action-planning and instruction-reasoning samples, along with 763k large-scale web data. Extensive experiments show that NaVid achieves state-of-the-art performance in simulation environments and the real world, demonstrating superior cross-dataset and Sim2Real transfer.
We thus believe our proposed VLM approach plans the next step for not only the navigation agents but also this research field.

\end{abstract}

\IEEEpeerreviewmaketitle

\section{Introduction}

As a fundamental task of Embodied AI, vision-and-language navigation (VLN) \cite{gu2022vision,park2023visual} requires the agents to navigate in diverse and especially unseen environments following free-form linguistic instructions. 
VLN requires robots to understand complex and diverse visual observations and meanwhile interpret fine-grained instructions~\cite{chen2019touchdown,vasudevan2021talk2nav}, ``\textit{go up the stairs and stop in the doorway}'', and thus maintains to be a challenging task. To address this challenging task, a large volume of research~\cite{shah2022lmnav,chen2021hamt,wang2023scaling,tsai2023multimodal,pan2023langnav,krantz2023iterative, an2023etpnav} in this field is launched in the simplified setting, \ieno, decision-making in discrete environments (\egno, R2R~\cite{krantz_vlnce_2020} in MP3D simulator~\cite{chang2017matterport3d}). Specifically, the real environments are abstracted as connectivity graphs, and the navigation is cast as the teleportation over a waypoint set on such graphs.
Despite these methods are rapidly evolving and yielding impressive results~\cite{shah2022lmnav, zhou2023navgpt,long2023discuss, wang2023scaling}, the discretized environment setting introduces additional challenges such as the need for a landmark graph~\cite{krantz2020beyond, krantz2021waypoint} and a local model to navigate between landmarks~\cite{shah2021ving, shah2023gnm, sethian1999fast}.


Toward more realistic and straightforward modelling, navigation in continuous environments, \egno, R2R-CE, RxR-CE, has garnered increasing attention. Considerable excellent research efforts are devoted to reducing the Sim-to-Real gaps \cite{krantz2021waypoint,hong2022bridging,xu2023vision,anderson2021sim}. However, they still face severe challenges in generalization due to data scarcity and domain gaps in their model inputs including RGBD, odometer data, or maps. The generalization issue presents a critical yet under-explored challenge in large-scale real-world deployment, encompassing transitions from seen scenes to novel environments and from simulation to real worlds (Sim-to-Real). The recent prosperity of large Vision Language Models (VLMs) has shown unprecedented promises in a lot of research fields~\cite{yang2023dawn,li2023blip,an2022bevbert}.
In this paper, we explore whether large models can do the same in propelling generalizable VLN.

\begin{figure}[t]
\begin{center}
  \includegraphics[width=1 \linewidth]{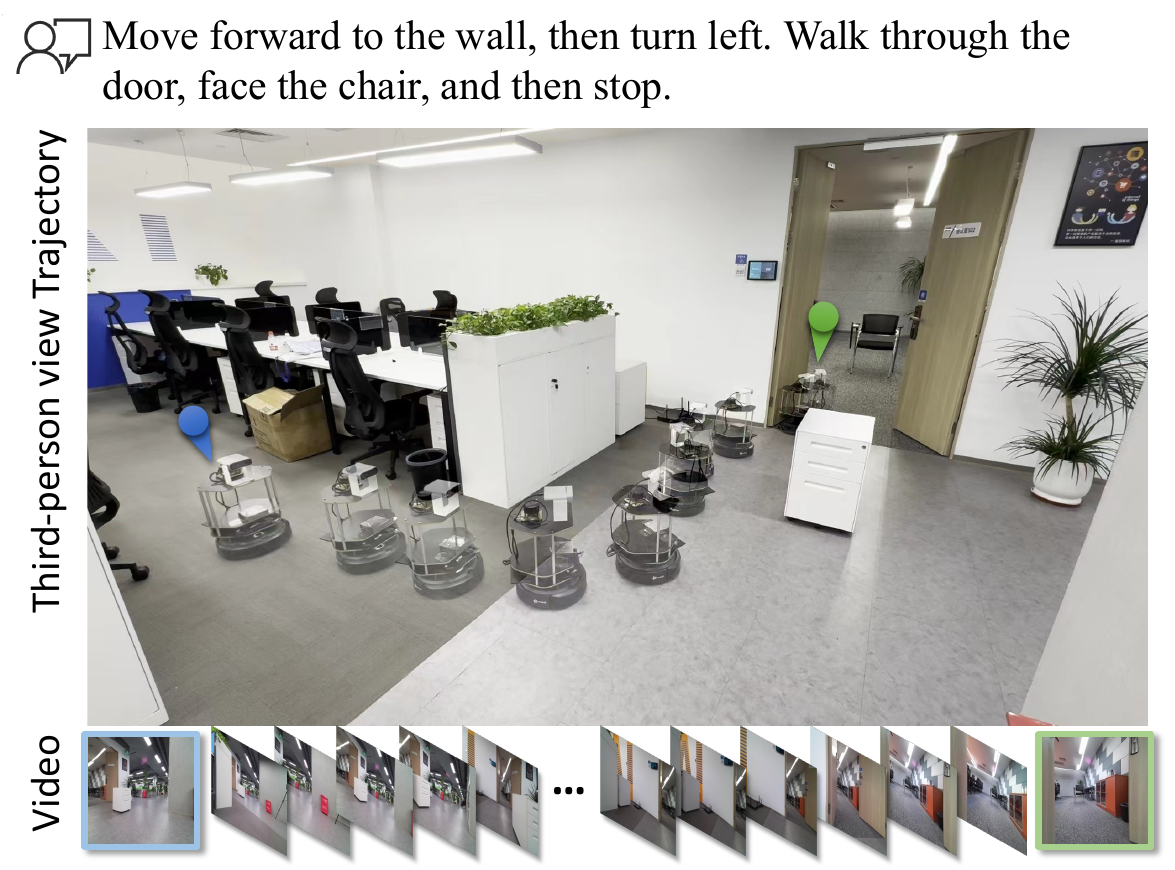}
\end{center}
   \caption{Real-world demo of our proposed video-based VLM, NaVid, for vision-and-language navigation. Given the human instruction, NaVid only takes online RGB video frames as input and outputs a language action for robotic execution. }
\label{fig:teaser}
\end{figure}

Large models have exhibited impressive generalization capacities in a broad range, covering AIGC \cite{rombach2022high,blattmann2023stable,zhang2023adding}, generalist chatbots \cite{achiam2023gpt}, autonomous driving~\cite{fu2024drive}, \etcno. They have also been shaping the future of embodied AI. RT-2~\cite{brohan2023rt} shows the promise of transferring web knowledge from VLMs to generalizable robotic manipulation. Large Language Models (LLMs) have been effective as the planners for VLN in discrete environments \cite{long2023discuss,pan2023langnav,zhou2023navgpt,chen2024mapgpt,qiao2023march}. The recent advancements in VLMs are ushering the VLN research into an exciting juncture. It is time to study \textit{whether VLMs can boost the generalization of VLN in continuous real-world environments.} 

In this paper, we make the first endeavour to capitalize on the power of foundational VLMs for generalizing VLN to the real world and propose a video VLM-based navigation agent, dubbed \ours. It solely relies on videos captured by a robot's monocular camera and humane-issued instructions as inputs for planning the next-step action in an end-to-end manner. We put the proposed \ours~into perspective by comparing it with three categories of models. 1) Compared with AGI models~\cite{yang2023dawn} or so-called navigation generalists~\cite{zheng2023towards} that can perform coarse navigation planning, \ours{} is a practical Vision-Language-Action (VLA) model that can infer executable actions with quantitative arguments, \egno, moving distances, and rotation degrees. 
This allows \ours{} to be deployed in the real world. 2) Compared to the VLN models that employ LLMs as planners, \ours~adopts a more realistic modeling for VLN. In particular, \ours~directly deduces low-level executable actions in continuous environments and encodes visual observations in video form, unlike previous LLM-based methods of modeling VLN in discrete spaces or encoding historical observations with textual descriptions~\cite{pan2023langnav, zhou2023navgpt, long2023discuss,chen2024mapgpt}. 3) Distinct from existing specialized VLN models, \ours~eliminates the reliance on odometer data, depths, and maps for action planning, thereby obviating the need for generalization challenges posed by odometer noises or the domain discrepancies in depth perception or navigation maps, making \ours~easy to be deployed.
To the best of our knowledge, the proposed \ours~is the first video-based VLM for VLN in continuous environments, achieving RGB-only navigation akin to human navigational behavior.

Our proposed \ours~employs a pre-trained vision encoder to encode visual observations and a pre-trained LLM to reason navigation actions. In this way, the general-purpose knowledge acquired in large-scale pretraining is transferred to VLN tasks, facilitating the learning and boosting the generalization. Drawing inspiration from the advanced video-based VLM, LLaMA-VID~\cite{li2023llama}, we represent each frame in robotic visual observations with two kinds of tokens.
The first kind consists of an instruction-queried token that extracts visual features specifically relevant to given instructions. The other kind comprises instruction-agnostic tokens that globally encode fine-grained visual information, where the token number determines the granularity of encoded features. 
The number of tokens for historical observations is allowed to differ from that for current observations. 
As such, in \ours, robotic historical trajectories are encoded as visual tokens in the video form, which provides a more informative and adaptive context compared to prior LLM-based VLN models that perform encoding in discrete spaces~\cite{chen2021hamt, chen2022duet} or using textual descriptions~\cite{pan2023langnav, zhou2023navgpt, long2023discuss,chen2024mapgpt}. Such video-based modeling imposes stringent constraints on the model inputs as it does not involve other information, \egno, depth, odometer data, or maps, except for the monocular video. When harnessed in the right manner, it advances in mitigating the generalization challenges resulting from odometer noises and domain discrepancies in depth perception or navigation maps of previous VLN works.


We conduct extensive experiments for the evaluation of our proposed \ours~in both simulated and real-world environments. Specifically, \ours~achieves SOTA-level performance on the VLN-CE R2R dataset and showcases a significant improvement on cross-dataset evaluation (R2R-RxR). Besides, our method demonstrates impressive robustness on Sim-to-Real deployment, achieving about $66\%$ success rate on 200 instructions across four diverse indoor scenes, leveraging only RGB videos as inputs.






\section{Related works}
\noindent \textbf{Vision-and-Language Navigation (VLN).}
A large endeavor of learning to navigate in unvisited environments following human instructions is established over discretized simulated scenes~\cite{anderson2018vision, anderson2020rxr, qi2020reverie, thomason2020cvdn}, where agents teleport between nodes on a pre-defined navigation graph by aligning language and visual observations for decision-making~\cite{ma2019self, wang2019reinforced, fried2018speaker, tan2019envdrop, ke2019tactical, fu2020counterfactual, qi2020object, hong2020graph}. 
Despite efficiency, directly transferring VLN models trained in discrete space to real-world robot applications is impractical. As a result, the more realistic VLN in continuous environments (VLN-CE) is proposed~\cite{krantz2020beyond, savva2019habitat}, allowing agents to navigate freely to any unobstructed space in the simulator either by predicting low-level controls~\cite{raychaudhuri2021law, irshad2021sasra, chen2021topological, georgakis2022cross, chen2022weakly} or selecting from navigable sub-goals estimated by waypoint predictors~\cite{hong2022bridging, krantz2021waypoint,krantz2022sim}.
Meanwhile, following the success of learning generic visual-linguistic representations from web-scale image-text pairs~\cite{chen2020uniter, li2020oscar, su2019vl, li2019visualbert, tan2019lxmert}, many VLN models benefit from large vision-language models~\cite{li2019robust,hong2020recurrent,chen2021hamt,chen2022duet} and VLN-specific pre-training~\cite{hao2020prevalent,majumdar2020improving,guhur2021airbert,wu2022cross,qiao2022hop}. Very recently, by scaling up navigation training data, the performance of VLN agents on the well-recognized R2R benchmark~\cite{krantz_vlnce_2020} is approaching humans~\cite{wang2023scaling}. This significant development suggests that the implementation of VLN techniques in real-world robotics is an increasingly viable and timely consideration.

\begin{figure*}[t]
\begin{center}
  \includegraphics[width=1 \linewidth]{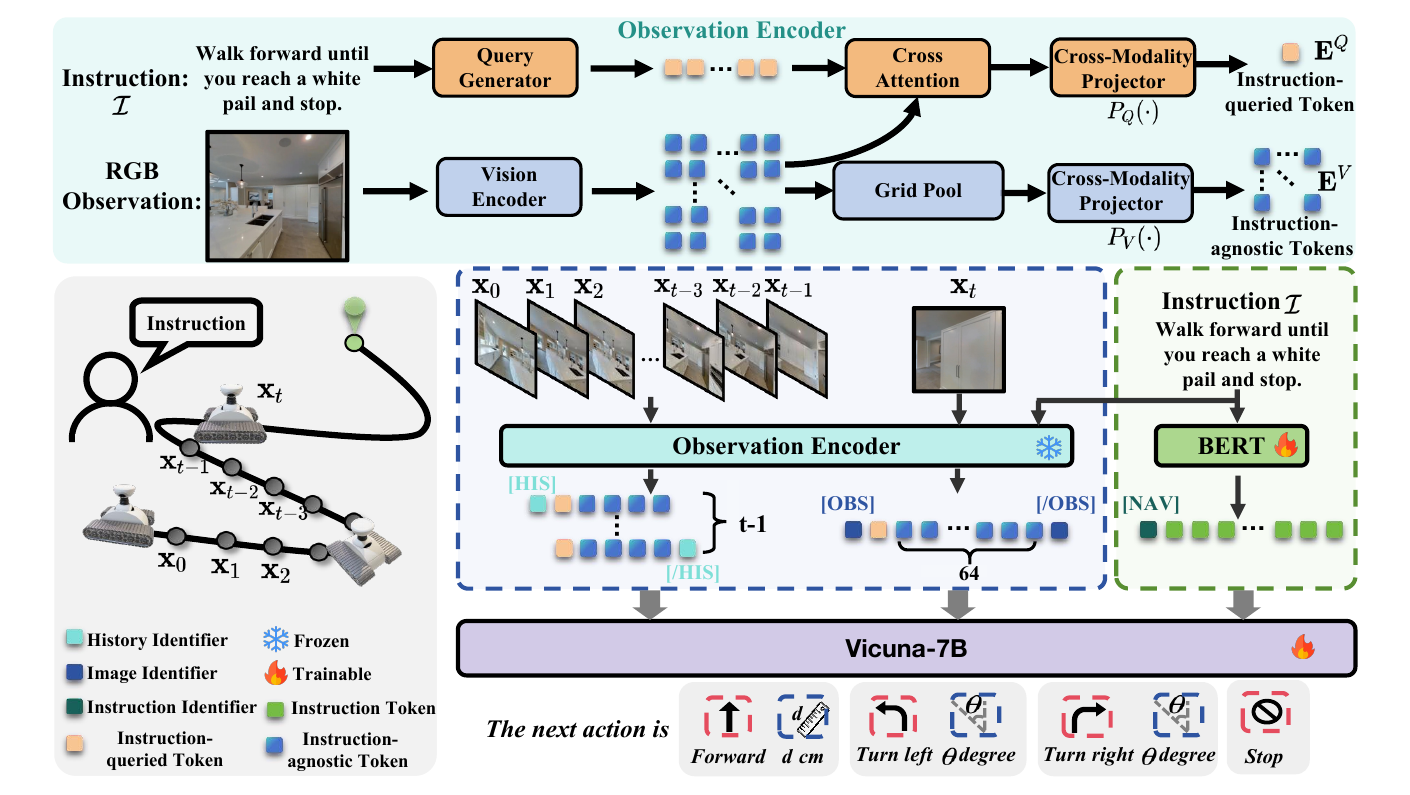}
\end{center}
   \caption{\textbf{The overview of \ours{}.} The inputs of \ours{} consist of the RGB frames from the online video observation $\{\mathbf{x}_0, \cdots, \mathbf{x}_t\}$ along with the human instruction $\cI$. For each frame, we use an observation encoder to extract the visual information with the instruction to obtain observation tokens, including, instruction-queried tokens (orange blocks) and instruction-agnostic tokens (blue blocks). At the current step $t$, the history frames $\{\mathbf{x}_0, \cdots, \mathbf{x}_{t-1}\}$ and the current frame $\mathbf{x}_t$ are encoded as observation tokens, with 4 and 64 instruction-agnostic tokens for history frames and current frames, respectively. Besides, our method obtains language tokens by a text encoder. Finally, split by the special tokens ~\texttt{[HIS]}, ~\texttt{[OBS]}, and ~\texttt{[NAV]}, we concatenate the observation tokens and language tokens and send the tokens to the Vicuna-7B then obtain the next-step action.
  }
\label{fig:backbone}
\end{figure*}

\noindent \textbf{Sim-to-Real Transfer for VLN.}
Despite great advances, existing VLN methods are predominantly built and evaluated in simulation, which largely overlooks the intricate and unpredictable nature of real-world conditions. 
Sim-to-real VLN transfer is an under-studied topic; until today, the only literature that systematically studies this problem is from Anderson et al.~\cite{anderson2020sim}, which justifies the performance gap (more than 50\% drop in success rate) due to action space and visual domain differences. In addition, we would like to highlight the challenge of generalizing to free-form language instructions - agents often fail to interpret different styles of instructions even when trained on millions of in-domain visual data~\cite{wang2023scaling,kamath2022marval}.
In light of this, many recent researches leverage the exceptional generalization abilities of the Large (Vision-)Language Models to facilitate VLN generalization. Either investigating the navigational reasoning capacity intrinsic to LLMs~\cite{zhou2023navgpt, pan2023langnav, zheng2023towards, long2023discuss, chen2024mapgpt, schumann2023velma, liang2023mo} or incorporating LLMs into navigation systems through modular designs that facilitate instruction parsing~\cite{chen20232} or through the injection of commonsense knowledge~\cite{qiao2023march}. 
We follow such a trend and further explore how to utilize a unified large model for low-level action prediction and its generalizability in real-world scenarios. This approach seeks not only to advance the state of VLN by leveraging the comprehensive understanding and versatile capabilities of LLMs~\cite{zhou2023navgpt, pan2023langnav, zheng2023towards, long2023discuss, chen2024mapgpt, qiao2023march} but also to bridge the gap between simulated environments and the multifaceted challenges presented by real-world applications.

\noindent \textbf{Large Models as Embodied Agents.}
Recently, researchers have been exploring the integration of large models into diverse embodied domains~\cite{driess2023palm,zhou2023navgpt,lin2023development,shah2023lm,song2023llm,schumann2023velma,huang2023inner}. PaLM-E \cite{driess2023palm}, for instance, suggests the incorporation of tokens from various modalities, alongside text tokens, into a large model. The model then generates high-level robotics instructions for tasks such as mobile manipulation, motion planning, and tabletop manipulation. Taking a step further, RT-2~\cite{brohan2023rt} generates low-level actions for robots, facilitating closed-loop control. GR-1 \cite{wu2023unleashing} introduces a GPT-style model~\cite{radford2018improving,zhou2023navgpt,long2023discuss} specifically designed for multi-task language-conditioned visual robot manipulation~\cite{zeng2021transporter}. This model predicts robot actions and future images based on language instructions, observed images, and robot states. RoboFlamingo \cite{li2023vision} proposes a vision-language manipulation framework that utilizes pre-trained vision-language models to formulate manipulation policies for robotics. It aims to offer a cost-effective and high-performance solution for robot manipulation~\cite{cui2021toward,suomalainen2022survey,zeng2021transporter,shridhar2022cliport}, allowing users to fine-tune their robots with large models. EMMA-LWM~\cite{Zhang2024LanguageGuidedWM} develops a world model for driving agents through verbal communication, showing convincing results in a digital gaming environment~\cite{Wang2021GroundingLT}.
Diving deeper into these works, this paper centers on another pivotal embodied domain: vision-and-language navigation, which necessitates robots to navigate in unseen environments following human instructions.

\section{Problem Formulation}
\label{sec:formulation}
The formulation of Vision-and-Language Navigation in Continuous Environments (VLN-CE) in this work is as follows: At the time $t$, given a natural language instruction $\cI$ consisting of $l$ words and a video observation $\cO_t$ comprising a sequence of frames $\{\mathbf{x}_0, \cdots, \mathbf{x}_t\}$, the agent is required to plan a low-level action $\mathbf{a}_{t+1} \in \cA$ for the next step. This action will take the agent to the next state and the agent will receive a new observation $\mathbf{x}_{t+1}$. Overall, we can formulate the decision-making as a Partially Observable Markov Decision Process (POMDP), denoted by $\{\mathbf{x}_0, \mathbf{a}_1, \mathbf{x}_1, \mathbf{a}_2, \cdots, \mathbf{x}_t\}$. In this work, the observation space $\cO$ corresponds to the videos captured with a monocular RGB camera without other additional data involved, and the action space incorporates qualitative action types alongside quantitative action arguments, also known as low-level actions \cite{krantz2020beyond} in this domain. This modeling enables a natural paradigm where observations are purely vision-based and readily obtainable, while actions are directly executable, akin to human navigational behavior.


\section{The Proposed \ours~Agent}
\label{sec:proposed_method}


With the formulation in Sec.\ref{sec:formulation}, we devise a video VLM-based navigation agent, named \ours. The \ours~is the first of its kind to transfer general knowledge of VLMs to a realistic VLN agent. We introduce its architecture in Sec. \ref{sec:model_architecture}, then elaborate on the detailed modeling for VLN inputs and outputs in Sec. \ref{sec:modeling}. The training strategy is detailed in Sec. \ref{sec:data_and_training}, and the implementation details are described in Sec. \ref{sec:implementation}. 

\subsection{Overall Architecture}
\label{sec:model_architecture}

We build \ours~on top of a general-purpose video-based VLM named LLaMA-VID~\cite{li2023llama}. For our proposed \ours, we inherit the main architecture of LLaMA-VID and incorporate the task-specific designs on top of it, to facilitate the transfer of general knowledge to VLN-CE to make its generalization challenges more readily solvable. 

As illustrated in Fig. \ref{fig:backbone}, \ours~consists of a vision encoder, a query generator, a LLM, and two cross-modality projectors. Given the observations up to time $t$, \ieno, a video sequence comprising $t$ frames, we encode this video to a sequence of tokens via the vision encoder (EVA-CLIP~\cite{sun2023eva} in implementation) and project them to a space aligned with language tokens. For brevity, we call the projected tokens as observation tokens. As common, the instructions are also tokenized as a set of tokens, called instruction tokens. We concatenate both observation tokens and instruction tokens and send them to the LLM to infer the VLN actions in linguistic form. Note that our work focuses on task-specific modeling rather than model architecture, as detailed in the following.

\subsection{The VLN-CE Modeling of \ours{}}
\label{sec:modeling}

\noindent \textbf{Observation encoding.} Given the captured monocular video up to time $t$, denoted by $\cO_t=\{\mathbf{x}_0. \cdots, \mathbf{x}_t\}$, we represent each frame with one instruction-queried visual token and several instruction-agnostic visual tokens. The instruction-queried tokens extract visual features specifically relevant to given instructions, and the instruction-agnostic tokens globally encode fine-grained visual information. For each frame $\mathbf{x}_t$, we first get its visual embedding $\mathbf{X}_t \in \mathbb{R}^{N_x \times C}$ with a vision encoder, where $N_x$ is the patch number ($N_x$ is set to 256) and $C$ is the embedding dimension. 

To get the instruction-queried tokens, we employ a Q-Former-based query generator to generate the instruction-aware query $\mathbf{Q}_t \in \mathbb{R}^{M \times C}$ with the query generator $G_Q$, where $M$ denotes the number of queries per frame and $C$ is the dimension of each query. The query generation can be formulated as:
\begin{equation}\label{eq:query_generation}
    \mathbf{Q}_t = G_Q(\mathbf{X}_t, \mathbf{I}),
\end{equation}
where $\mathbf{X}_t$ and $\mathbf{I}$ are the visual embeddings and the textual embeddings of the frame $\mathbf{x}_t$ and the instruction $\cI$. $G_Q(\cdot)$ is a Q-Former-based transformer to learn instruction-aware queries via a cross-modality interaction between $\mathbf{X}_t$ and $\mathbf{I}$ 
as that in ~\cite{li2023llama}. Similar to the operation in Q-Former \cite{li2023blip}, the instruction-queried tokens $\mathbf{E}^Q_t$ are obtained by a cross-attention between $\mathbf{X}_t$ and $\mathbf{Q}_t$, formulated as below:
\begin{equation}\label{eq:instruct_queried}
    \mathbf{E}^Q_t = P_Q(Pool(Softmax(\mathbf{Q}_t\mathbf{X}_t^T)\mathbf{X}_t)),
\end{equation}
where $P_Q(\cdot)$ represents the cross-modality projector for getting instruction-quired tokens, and $Pool(\cdot)$ is an averaging operation along the dimension of queries, making $\mathbf{E}^Q_t \in \mathbb{R}^{1\times C}$.

For the instruction-agnostic tokens, we directly perform a grid pooling operation and a cross-modality projection to get them, which can be formulated as:
\begin{equation}\label{eq:instruct_agnostic}
    \mathbf{E}^V_t = P_V(GridPool(\mathbf{X}_t)),
\end{equation}
where $GridPool(\cdot)$ is a grid pooling operation~\cite{li2023llama}, squeezing the tokens from $N_x$ to $N_v$, yielding $\mathbf{E}^V_t \in \mathbb{R}^{N_v\times C}$. A detailed description of the grid pooling can be found in the supplemental material. Representing each frame with two tokens as in LLaMA-VID \cite{li2023llama} does not meet the requirements of the VLN-CE task, as experimentally evidenced in the following. This is because LLaMA-VID is primarily designed for high-level question-answering tasks, whereas NaVid needs to plan executable actions for robots. Thus, we adopt the grid pooling here to enable the instruction-agnostic tokens to preserve sufficient geometry information so that the LLM in \ours~have enough contexts for reasoning the quantitative arguments of robotic actions.

For VLN-CE, the current frame serves as the primary basis for navigation action reasoning, while historical frames provide important contexts for tracing navigation progress. Considering their different requirements for preserving geometry information, we adopt a varying number of instruction-agnostic tokens in encoding the historical frames and the current frame. In this work, unless otherwise specified, we set the number of instruction-agnostic tokens to 64 for the current frame while 4 for each historical frame. This not only facilitates the learning but also improves the efficiency. To further facilitate the training of \ours, we explicitly distinguish different types of information with special tokens before sending them to the LLM within \ours.
Specifically, we adopt special tokens \texttt{<HIS>}, \texttt{</HIS>}, and \texttt{<OBS>}, \texttt{</OBS>} to demarcate tokens encoded from historical frames and the current frame, respectively. Here, \texttt{<HIS>} and \texttt{<OBS>} signify the beginning of the corresponding token sequences while \texttt{</HIS>} and \texttt{</OBS>} mark their ends. Additionally, we use another special token \texttt{<NAV>} to prompt the LLM to begin to process the textual tokens of instructions and output the robotic actions in linguistic form. As a result, the input of \ours~can be summarized as below. 
\begin{tcolorbox}
\textbf{Input:} $<\!HIS\!>\{historical\_frames\}<\!/HIS\!><\!OBS\!>\{current\_frame\}<\!/OBS\!>)<\!NAV\!>\{instruction\_content\}$\\
\textbf{Output:} $\{answer\_content\}$
\end{tcolorbox}
In this format, $\{historical\_frames\}$, $\{current\_frame\}$, $\{instruction\_content\}$ and $\{answer\_content\}$ are the placeholders for the tokens of historical frames, current frame, instruction, and reasoned actions, respectively.

\noindent \textbf{Action planning.} \ours~plans the next-step action for VLN-CE in linguistic form. Each action of its outputs consists of two variables aligned with the setting of VLN-CE. One of them is the action type, chosen from a discrete set $\{\texttt{FORWARD}, \texttt{TURN-LEFT}, \texttt{TURN-RIGHT}, \texttt{STOP}\}$. The other is the quantitative arguments corresponding to different action types. For $\texttt{FORWARD}$, \ours~further infers the specific moving distance. For $\texttt{TURN-LEFT}$ and $\texttt{TURN-RIGHT}$, \ours~also predicts specific rotation degrees. A regular expression parser \cite{kearns1991extending} is employed to extract the action types and arguments for model evaluation and real-world deployment.

\subsection{The Training of \ours{}}
\label{sec:data_and_training}


The available navigational simulation data are still limited in their diversity, authenticity, and scale. We design a hybrid training strategy to maximize the utilization of these data in enabling \ours~to generalize as effectively as possible to novel scenes or the real world. To this end, two key approaches are proposed in the hybrid training strategy. One is to collect non-oracle navigation trajectories and incorporate them into the training loop. The other is to design auxiliary tasks to enhance the capabilities of \ours~on navigation scene understanding and instruction following. We elaborate separately below.

\begin{figure}[t]
\begin{center}
  \includegraphics[width=1 \linewidth]{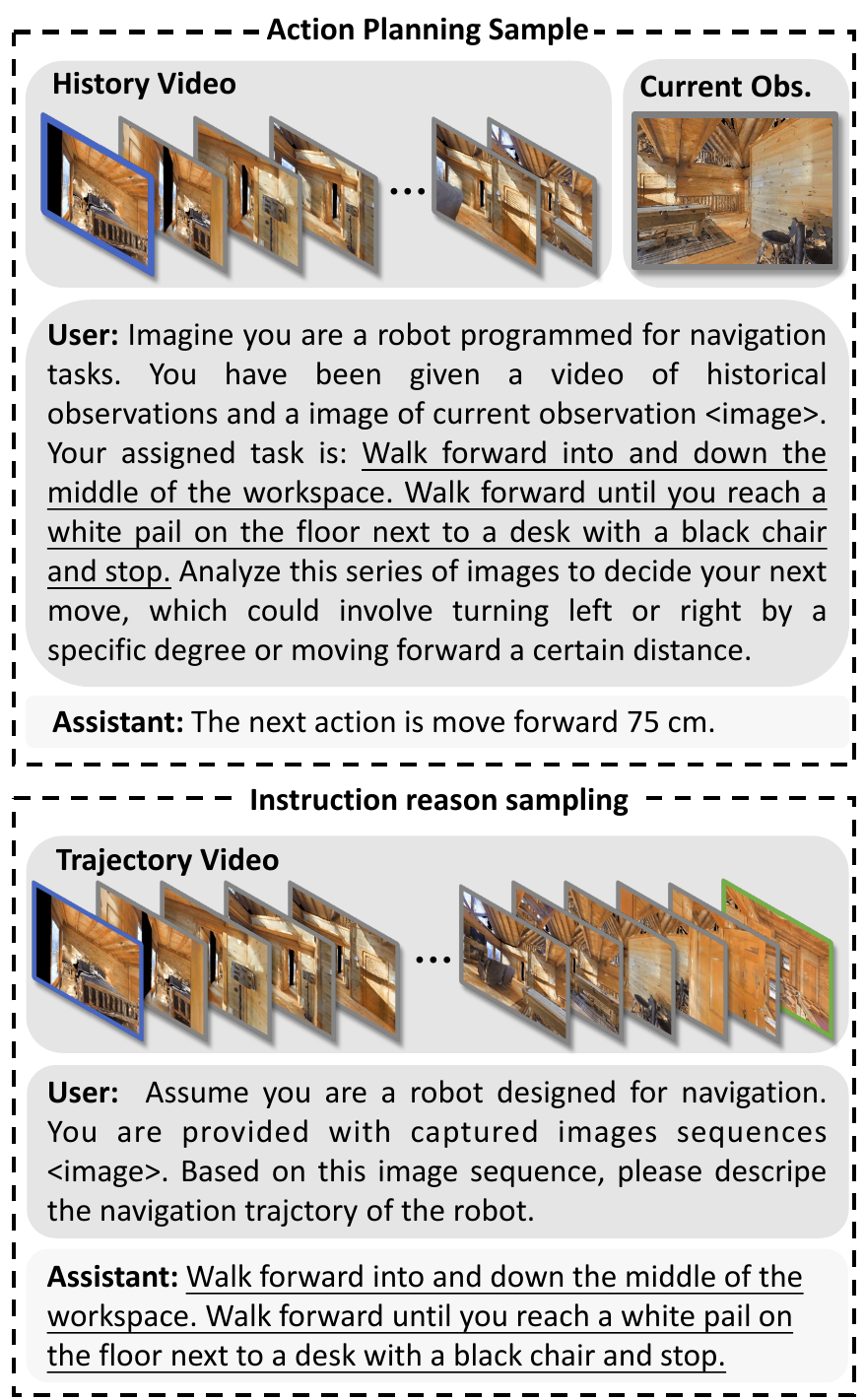}
\end{center}
   \caption{Examples of navigation data, comprising the action planning sample (above) and the instruction reasoning sample (below).}
\label{fig:data_and_training}
\end{figure}

\textbf{Non-oracle navigation trajectories collection.} Inspired by the Dagger technique \cite{ross2011dagger}, we collect non-oracle navigation trajectories and incorporate them into the training of \ours. Without this approach, our \ours~would only be exposed to oracle navigation trajectories during training, which diverges from practical application conditions and diminishes the robustness of the learned navigation strategy. To achieve this, we first collect oracle navigation trajectories, including monocular video observations, instructions, and robotic actions, from the VLN-CE R2R dataset. Specifically, we gather data from 61 MP3D indoor scenes \cite{chang2017matterport3d}, containing 320k step-wise samples in total. Then, we train \ours~on these oracle trajectory data and deploy the obtained agent in VLN-CE environments for further collecting the non-oracle navigation trajectories. As a result, we get another 180k step-wise samples. The samples from both oracle and non-oracle trajectories are combined for the final training of \ours, as shown in the above of Fig. \ref{fig:data_and_training}.

\textbf{Co-training of VLN-CE and auxiliary tasks.} As a navigation agent, precisely understanding environments and following given instructions are two indispensable capabilities besides planning navigation actions. To facilitate agent learning, we combine the VLN-CE action planning together with two auxiliary tasks in a co-training paradigm. 
For environmental understanding, we have designed an auxiliary task called instruction reasoning. Given a video-based navigation trajectory, \ours~ is required to deduce the corresponding instructions for that trajectory.
This auxiliary task can be easily implemented with a shared data organization format introduced in Sec. \ref{sec:modeling}, in which the $\{instruction\_content\}$ and $\{answer\_content\}$ can be instantiated as the prompts of requesting the descriptions of robotic navigation trajectories and the human-labeled instructions provided in the dataset. The instruction reasoning auxiliary task contains 10k trajectories, an example is provided in the below of Fig. \ref{fig:data_and_training}. Moreover, for the enhancement of instruction following and the anti-forgetting of general knowledge acquired in pre-training, we incorporate the video-based question-answering samples into our co-training as well. Details can be found in \cite{li2023llama}. For brevity, we do not elaborate on them here.

\subsection{Implementation details}
\label{sec:implementation}

\textbf{Training configurations.} NaVid is trained on a cluster server with 24 NVIDIA A100 GPUs for approximately 28 hours, totaling 672 GPU hours. For video-caption data, we sample frames at 1 FPS, to remove redundant information between consecutive frames. We keep all the frames for navigation-action data, mostly smaller than 300 frames. During training, all modules including EVA-CLIP~\cite{sun2023eva}, QFormer~\cite{dai2023instructblip}, BERT~\cite{devlin2018bert}, and Vicuna-7B~\cite{chiang2023vicuna} are loaded with default pre-trained weight. Following the strategy in \cite{li2023llama}, we only optimize the trainable parameters of LLaMA and text encoder for only 1 epoch.

\textbf{Evaluation configurations.} 
After~\ours{} predicts language actions, we leverage regular expressions matching~\cite{kearns1991extending} for expecting valid actions.
We find this simple algorithm achieves a $100\%$ success rate for obtaining valid actions under the VLN-CE val-unseen R2R evaluation.
For real-world navigation, we use a remote server to run NaVid to receive observations (along with text instructions) and command a local robot to execute the predicted actions. During navigation, the agent requires about 1.2 to 1.5 seconds to output one action per frame. This could be improved by using acceleration techniques, such as quantization~\cite{liu2023llava, liu2023improved}.


\section{Experiments}

\subsection{Experiment Setup}
\label{sec:exp-setup}

\textbf{Simulated environments.} We evaluate our method on the VLN-CE benchmarks, which provide continuous environments for executing low-level actions on reconstructed photorealistic indoor scenes~\cite{krantz2020beyond}. We consider R2R~\cite{krantz_vlnce_2020} and RxR~\cite{anderson2020rxr} in VLN-CE, the two most recognized benchmarks of VLN. For a fair comparison, all methods are trained on the 10,819 R2R train split, and evaluated on the 1,839 R2R val-unseen split and 1,517 RxR val-unseen split to evaluate cross-split and cross-datasets performance, respectively.

\textbf{Real-world environments.} To evaluate the performance of our method in real-world environments, we follow the experiment setting of~\cite{anderson2020sim, xu2023vision} and design comprehensive experiments that include different indoor scenes and different difficulty instructions. We select four diverse indoor scenes including \texttt{Meeting\_room}, \texttt{Office}, \texttt{Lab}, and \texttt{Lounge}. For instruction, we design two types of instruction: 1) Simple landmark instruction following task, which requires the agent to understand the semantics of the target and move to the relative location of the target. 2) Complex Composite instruction following Task, which requires the agent to complete 2-5 combination instructions of landmark following task. For each selected scene, we design 25 Simple landmark instructions and 25 complex composite instructions, leading to a total of 200 instruction-following cases. 
%

We conduct all real-world experiments using a Turtlebot4\footnote{Turtlebot4 overview: \href{https://clearpathrobotics.com/turtlebot-4/}{https://clearpathrobotics.com/turtlebot-4/}.} equipped with a Kinect DK camera for capturing both depth and RGB images. For baselines that require odometry, we use a built-in RPLIDAR A1M8 Lidar of the turtlbeot4 and leverage the toolbox of Nav2~\cite{macenski2020marathon} for localization and mapping. The calibration of the Lidar and camera follows the previous work~\cite{zhang2004extrinsic}.
Please refer to the supplemental material for details. Note that there exist advanced tracking and localization systems~\cite{liu2023efficient,zhang2021rosefusion, zhang2022asro} which may further improve the performance.

\textbf{Metrics.} We follow the standard VLN evaluation metrics~\cite{anderson2018vision, krantz_vlnce_2020, ku2020room, kuang2024openfmnav} to evaluate the navigation performance, including success rate (SR), oracle success rate (OS), success weighted by path length (SPL)~\cite{anderson2018spl}, trajectory length (TL), and navigation error from goal (NE). Among all, SPL is the primary metric as it reflects both the accuracy and efficiency of navigation~\cite{anderson2018evaluation}.
Note that an episode is considered successful if the agent calls the STOP action within 3 m of the goal in the VLN-CE and 1.5 m in real-world environments. More details about evaluation metrics can be found in the supplemental material.

\textbf{Baselines.} For a fair comparison with \ours{}, we compare the performance of methods that directly predict low-level action primitives in the VLN-CE environments:
\begin{itemize}[noitemsep,nolistsep,leftmargin=*]
    \item Seq2Seq~\cite{krantz2020beyond} is a simple sequence-to-sequence baseline that uses a recurrent policy to predict
the action directly from the RGBD observations. RGB-Seq2Seq indicates the use of RGB observations only.
    \item CMA~\cite{krantz2020beyond} utilizes cross-model attention between instruction and RGBD-observations to predict action.  RGB-CMA indicates the use of RGB observations only.
    \item WS-MGMap~\cite{chen2022weakly} leverages a multi-granularity map, which contains object geometry, texture, and semantic information.
\end{itemize}

\subsection{Comparison on Simulated Environment}

\begin{table}[!t]
\centering
\caption{\jiazhao{Comparing on VLN-CE R2R Val-Unseen. $^*$: Methods use high-level action space. $\dag$: Methods use the same waypoint predictor proposed in~\cite{hong2022bridging}. $\ddag$: Methods use additional visual data than MP3D scenes~\cite{chang2017matterport3d}.}
}
\resizebox{\linewidth}{!}{
\scalebox{1}{
\setlength{\tabcolsep}{0.8mm}{
\begin{tabular}{l|ccccccccc}
\hline
\multirow{2}{*}{} & \multicolumn{4}{c|}{Observation}                                        & \multicolumn{5}{c}{VLN-CE R2R Val-Unseen}                                                                               \\ \cline{2-10} 
                  & Pan.       & S.RGB     & Depth      & \multicolumn{1}{c|}{Odo.}       & TL  & \textbf{NE}$\downarrow$& \textbf{OS}$\uparrow$    & \textbf{SR}$\uparrow$    & \textbf{SPL}$\uparrow$   \\ \hline \hline
HPN+DN$^*$~\cite{krantz2021waypoint}           & \checkmark &            &    \checkmark        & \multicolumn{1}{c|}{\checkmark} & 7.62          & 6.31                                                    & 40.0          & 36.0          & 34.0          \\
CMA$^*$$\dag$~\cite{hong2022bridging}           & \checkmark &            &   \checkmark         & \multicolumn{1}{c|}{\checkmark} & 10.90         & 6.20                                                    & 52.0          & 41.0          & 36.0         \\
\vlnbert$^*$$\dag$~\cite{hong2022bridging}      & \checkmark &            &   \checkmark         & \multicolumn{1}{c|}{\checkmark} & 12.23         & 5.74                                                    & 53.0          & 44.0          & 39.0         \\
Sim2Sim$^*$~\cite{krantz2022sim}      & \checkmark &            &   \checkmark         & \multicolumn{1}{c|}{\checkmark} & 10.69         & 6.07                                                    & 52.0          & 43.0          & 36.0         \\
GridMM$^*$$\dag$~\cite{wang2023gridmm}      & \checkmark &            &   \checkmark         & \multicolumn{1}{c|}{\checkmark} & 13.36         & 5.11                                                    & 61.0          & 49.0          & 41.0         \\
HAMT$^*$$\dag$$\ddag$~\cite{wang2023scaling}     & \checkmark &            &   \checkmark         & \multicolumn{1}{c|}{\checkmark} & --         & 4.80                                                    & --          & 55.0          & 51.0         \\
ETPNav$^*$~\cite{an2023etpnav}     & \checkmark &            &   \checkmark         & \multicolumn{1}{c|}{\checkmark} & 11.99& 4.71          & 65.0          & 57.0     & 49.0       \\
\hline
AG-CMTP~\cite{chen2021topological}           & \checkmark &            &   \checkmark         & \multicolumn{1}{c|}{\checkmark} & --             & 7.90                                                    & 39.2          & 23.1          & 19.1          \\
R2R-CMTP~\cite{chen2021topological}         & \checkmark &            &    \checkmark        & \multicolumn{1}{c|}{\checkmark} & --             & 7.90                                                    & 38.0          & 26.4          & 22.7          \\
LAW~\cite{raychaudhuri2021law}               &            & \checkmark & \checkmark & \multicolumn{1}{c|}{\checkmark} & 8.89          & 6.83                                                    & 44.0          & 35.0          & 31.0          \\
CM2~\cite{georgakis2022cross}               &            & \checkmark & \checkmark & \multicolumn{1}{c|}{\checkmark} & 11.54         & 7.02                                                    & 41.5          & 34.3          & 27.6          \\
WS-MGMap~\cite{chen2022weakly}          &            & \checkmark & \checkmark & \multicolumn{1}{c|}{\checkmark} & 10.00         & 6.28                                                    & 47.6          & \textbf{38.9} & 34.3 \\
Seq2Seq~\cite{krantz2020beyond}           &            & \checkmark & \checkmark & \multicolumn{1}{c|}{} & 9.30          & 7.77                                                    & 37.0          & 25.0          & 22.0          \\
CMA~\cite{krantz2020beyond}               &            & \checkmark & \checkmark & \multicolumn{1}{c|}{} & 8.64          & 7.37                                                    & 40.0          & 32.0          & 30.0          \\
RGB-Seq2Seq              &            & \checkmark &            & \multicolumn{1}{c|}{}           & 4.86 &  10.1                                           & 8.10 & 0.00          & 0.00 \\
RGB-CMA              &            & \checkmark &            & \multicolumn{1}{c|}{}           & 6.28 & 9.55                                           & 10.8 & 5.00          & 4.43 \\
\textbf{Ours}              &            & \checkmark &            & \multicolumn{1}{c|}{}           & 7.63 & \textbf{5.47}                                           & \textbf{49.1} & 37.4          & \textbf{35.9}          \\ \hline
\end{tabular}
}
}}
\label{tab:comp-vlnce-r2r}
\end{table}

\textbf{VLN-CE R2R.} 
We first conduct experiments on the VLN-CE R2R Val-Unseen dataset to evaluate the cross-split generalizability of methods. The results can be found in Table \ref{tab:comp-vlnce-r2r}. Here we mark the specific observations required by the methods, whereas our method only uses RGB observations.
Compared to the related VLN-CE setting (low-level action space), our method achieves SOTA performance in terms of SPL, NE, OS, and comparable performance on SR, without using depth and odometry information. 
For the methods Seq2Seq-RGB and CMA-RGB that share the same setting, our method significantly outperforms these methods by improving the SR from $5.00\%$ to $37.4\%$, and SPL from $4.43\%$ to $35.9\%$. Our method outperforms the SOTA method WS-MGMap by $1.6\%$ regarding SPL. The results demonstrate the effectiveness of our method by showing SOTA-level performance with only RGB observations.

\begin{figure}[t]
\begin{center}
  \includegraphics[width=0.95 \linewidth]{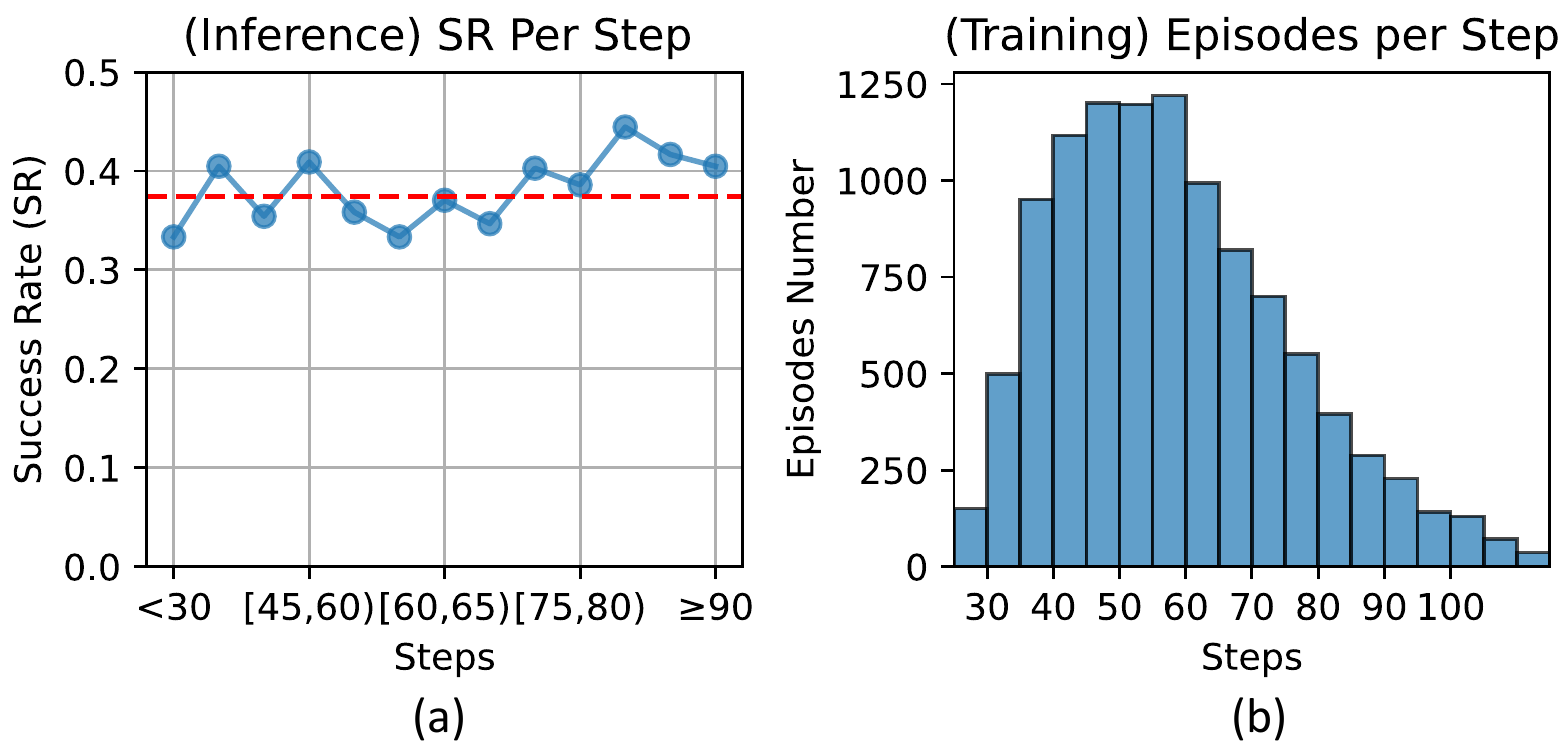}
\end{center}
   \caption{(a) Success Rate of NaVid on different steps during inference. The red dot line indicates the average success rate. (b) The episodes number on different steps during training. The 'Steps' of the x-axis indicate the oracle actions required by the instructions.}
\label{fig:long-horizion}
\vspace{-3mm}
\end{figure}

To explore the challenges associated with longer memory requirements (more steps) in NaVid, we report the performance on different horizon instructions, where instructions typically necessitate 30 to 90 steps to complete, spanning approximately 5 to 20 meters. The results are illustrated in Fig~\ref{fig:long-horizion} (a), showing that our method maintains consistent performance across a range from short (lower than 30 steps) to long distances (more than 90 steps). Additionally, we plot the number of episodes per step during training in Fig~\ref{fig:long-horizion} (b). Although about $5.91\%$ of episodes exceeded 90 steps, our method still performs comparably to the majority of episodes, which range from 35 to 75 steps (about $74.3\%$ of episodes). This robust performance may be attributed to our strategy of co-tuning navigation data with web-scale video data, which includes sequences of up to 300 frames, thus enabling our method to process a longer historical context than is typically required by most VLN tasks.

\begin{table}[!t]\centering
\caption{
\jiazhao{Comparing cross-dataset performance on VLN-CE RxR Val-Unseen. Note that, $A^2$Nav is a zero-shot method that leverages GPT as the planner.}
}
\resizebox{\linewidth}{!}{
\setlength{\tabcolsep}{1.1mm}{
\begin{tabular}{l|cccccccc}
\hline
\multirow{2}{*}{} & \multicolumn{3}{c}{Observation}                           & \multicolumn{5}{c}{VLN-CE RxR Val-Unseen}                                                \\ \cline{2-9} 
                  & S.RGB     & Depth      & \multicolumn{1}{c|}{Odo.}       & TL  & \textbf{NE}$\downarrow$  & \textbf{OS}$\uparrow$    & \textbf{SR}$\uparrow$    & \textbf{SPL}$\uparrow$   \\ \hline \hline
LAW~\cite{raychaudhuri2021law}               & \checkmark & \checkmark & \multicolumn{1}{c|}{\checkmark} & 4.01 & 10.87         & 21.0          & 8.0           & 8.0           \\
CM2~\cite{georgakis2022cross}         & \checkmark & \checkmark & \multicolumn{1}{c|}{\checkmark} & 12.29         & 8.98          & 25.3          & 14.4         & 9.2          \\
WS-MGMap~\cite{chen2022weakly}          & \checkmark & \checkmark & \multicolumn{1}{c|}{\checkmark} & 10.80         & 9.83          & 29.8          & 15.0          & 12.1          \\
Seq2Seq~\cite{krantz2020beyond}              & \checkmark &  \checkmark          & \multicolumn{1}{c|}{}           & 1.16          & 11.8 & 5.02 & 3.51 & 3.43 \\
CMA~\cite{krantz2020beyond}              & \checkmark & \checkmark           & \multicolumn{1}{c|}{}           & 5.09         & 11.7 & 10.7 & 4.41 & 2.47 \\
RGB-Seq2Seq              & \checkmark &            & \multicolumn{1}{c|}{}           & 4.43          & 11.2 & 12.2 & 0.0 & 0.0 \\
RGB-CMA              & \checkmark &            & \multicolumn{1}{c|}{}           & 13.56          & 9.55 & 14.8 & 0.0 & 0.0 \\
$A^2$Nav~\cite{chen20232}            & \checkmark &            & \multicolumn{1}{c|}{}           & --          & -- & -- & 16.8 & 6.3 \\
\textbf{Ours}              & \checkmark &            & \multicolumn{1}{c|}{}           & 10.59          & \textbf{8.41} & \textbf{34.5} & \textbf{23.8} & \textbf{21.2} \\ \hline
\end{tabular}
}
}
\label{tab:comp-rxr}
\end{table} 

\textbf{VLN-CE RxR.} 
To evaluate the cross-dataset performance of methods, we train the models on R2R trajectory-instruction samples and compare their zero-shot performance on RxR Val-Unseen data split. 
RxR dataset contains instructions of finer granularity that describe rich landmarks and longer trajectories, which establishes a large language and visual gap that is suitable for evaluating the generalization potential of the methods.
As shown in Table~\ref{tab:comp-rxr}, we find that our method outperforms existing methods by a large margin in terms of NE, OS, SR, and SPL. Note that the trajectory length (TL) of Seq2Seq is extremely small because of its poor instruction-following ability ($3.51\%$ SR), leading by early stopping. Compared with $A^2$Nav~\cite{chen20232}, the current SOTA method in zero-shot VLN-CE RxR, our method achieves $41.7\%$ (from 16.8 to 23.8) and $236.5\%$ (from 6.3 to 21.2) better results on SR and SPL. Besides, the Seq2Seq-RGB and CMA-RGB baselines, which share the same setting, fail on all instructions, demonstrating the challenges of RGB-only methods. This proves the generalization potential of our method by incorporating the video-based large vision-language model into VLN for addressing free-form language navigation tasks.


\begin{table}[!t]\centering
\caption{
\jiazhao{Comparison of LLMs on 100 sampled VLN-CE R2R Val-Unseen split. Here, the '-' indicates the corresponding method does not predict valid language actions for execution.}
}
\scalebox{1}{
\setlength{\tabcolsep}{2.0mm}{
\begin{tabular}{l|ccccc}
\hline
\multirow{2}{*}{} & \multicolumn{5}{c}{VLN-CE R2R Val-Unseen Sub-Split} \\ \cline{2-6} 
 & TL & \textbf{NE}$\downarrow$ & \textbf{OS}$\uparrow$ & \textbf{SR}$\uparrow$ & \textbf{SPL}$\uparrow$ \\ \hline \hline
GPT-4V \cite{yang2023dawn} & 5.79 & 11.4 & 10.00 & 5.00  & 3.11 \\
Emu \cite{sun2023generative} & - & - & - & - & - \\
LLaVA \cite{liu2023improved} & - & - & - & - & - \\
LLaMA-VID \cite{li2023llama} & - & - & - & - & - \\ \hline
LLaVA-Nav & 5.05 & 7.82 & 14.0 & 10.0 & 9.43 \\
LLaMA-VID-Nav & 10.7 & 8.73 & 40.0 & 29.0 & 27.5 \\
\textbf{Ours} & 8.02 & \textbf{5.52} & \textbf{45.0} & \textbf{38.0} & \textbf{35.4} \\ \hline
\end{tabular}
    }
}
\label{tab:comp-llm}
\end{table} 

\textbf{Comparison with Large foundation models.} We then want to evaluate the performance of variant mainstream large foundation models on the vision language navigation task. Here, GPT-4V \cite{yang2023dawn} and Emu \cite{sun2023generative} support multi-conversation with image inputs, so in each iteration of conversation we feed a new observed RGB image. LLaVA \cite{liu2023improved} is a sing-conversation image model, so we leverage an observation-to-history technique to encode history. LLaMA-VID~ \cite{li2023llama} is able to answer questions based on the video inputs, which can be replaced by the navigation observations. 
For the models LLaVA and LLaMA-VID, which have published their training datasets and code, we co-tune these models with our proposed navigation data, resulting in the modified models named LLaVA-Nav and LLaMA-VID-Nav, respectively.
All baselines are carefully tuned and promoted, and the details can be found in supplemental material.

Considering the extreme computation cost such as promoting history in GPT-4V and LLaVA, we randomly sample 100 episodes as a sub-split of VLN-CE R2R Val-Unseen sub-split (the ID of sampled episodes can be found in the supplemental material for reproducing). The results can be found in Table~\ref{tab:comp-llm}. We find that the methods without training on proposed navigation data suffer a very poor performance. Especially, Emu, LLaVA, and LLaMA-VID frequently output action-irrelevant answers, such as environment description or navigation comments, making the agent very hard to navigate. A more powerful model GPT-4V, with a carefully in-context prompt~\cite{zhou2023navgpt, hu2023look}, can relatively stable output valid actions, showcasing a lower performance at $5\%$ SR. After tuning with the proposed navigation data, we found significant improvements in the LLaVA-Nav and LLaMA-VID-Nav, which demonstrate the importance of the collected navigation data.
Besides, training on navigation data leads to better instruction-following ability, making the valid actions answer ratio from $0\%$ to $91.3\%$ for LLaMA-VID-Nav.

\begin{table}\centering
\caption{
\jiazhao{Comparison of different alternatives for modelling past trajectories.}
}
\scalebox{1}{
\setlength{\tabcolsep}{2.5mm}{
\begin{tabular}{l|ccccc}
\hline
 & \multicolumn{5}{c}{VLN-CE R2R Val-Unseen} \\ \cline{2-6} 
Method & TL & NE↓ & OS↑ & SR↑ & SPL↑ \\ \hline \hline
(A) Text-based & 1.45 & 8.82 & 0.0 & 0.0 & 0.0 \\
(B) Map-text & 2.76 & 7.12 & 9.51 & 9.13 & 8.97 \\
(C) Ego-view-text & 5.05 & 8.85 & 35.5 & 23.5 & 20.8 \\
Video-based (NaVid) & 7.63 & \textbf{5.47} & \textbf{49.1} & \textbf{37.4} & \textbf{35.9} \\ \hline
\end{tabular}
    }
}
\label{tab:comp-representation}
\end{table}

\textbf{Comparison with different navigation history representations.}
We compare our proposed video-based modeling in \ours~with other alternatives for representing past trajectories. We adjust the special tokens mechanism for identifying different representations while preserving other design elements unchanged to ensure a fair comparison. We have meticulously implemented these alternatives and collected appropriate navigation and caption data for training. Their detailed implementations are described below:

\begin{itemize}[noitemsep,nolistsep,leftmargin=*]
    \item \textbf{(A) Text-based}: We use text to describe the navigation history as inputs (including the current frame). By following NavGPT \cite{zhou2023navgpt}, we use a visual foundation model (LLaVA \cite{liu2023improved}) to describe the environments of each frame and combine per-frame text observation with GPT-4 \cite{achiam2023gpt} every 10 frames. Note that, using GPT-4V \cite{yang2023dawn} to replace LLaVA may have better performance but will lead to forbidden costs. The model is co-tuned with the text-QA dataset in LLaVA  \cite{liu2023improved}.
    \item \textbf{(B) Map-text}: We add the current top-down 2D map along with textual navigation history information (A) as inputs. The top-down 2D maps are obtained from the Habitat simulator \cite{savva2019habitat} and we directly use the default style. The top-down maps include an arrow indicating the location and orientation of the robots, and also use grey color to indicate navigatable areas. The top-down map is encoded as 256 tokens. The model is co-tuned with text- and image-QA datasets used in LLaMA-VID \cite{li2023llama} and LLaVA \cite{liu2023improved}.
    \item  \textbf{(C) Ego-view-text}: This model leverages the current egocentric image along with textual navigation history information (A) as inputs. The egocentric images are obtained directly from the Habitat simulator. The egocentric image view is encoded as 256 tokens.  The model is co-tuned with text- and image-QA datasets used in LLaMA-VID \cite{li2023llama} and LLaVA \cite{liu2023improved}.
\end{itemize}

We can find, for VLN tasks, modeling past trajectories as texts or 2D maps are both clearly inferior to modeling them in a video form, with SPL dropping from 35.9\% to 20.8\%/ 8.97\%.
This demonstrates the necessity and superiority of our proposed video-based modeling for the past trajectories. 
We believe the reason is that modeling past trajectories as texts or 2D maps significantly compresses the rich visual information, which increases the difficulty of understanding the trajectory history.
Furthermore, the inference time for textual history representation averages approximately 2.7 seconds per action step prediction. This duration is roughly double that of the video-based representation. The extended inference time comes from the necessity of an additional LLaVA query for image captioning and a GPT query for summarizing the navigation history.

\begin{table}[t] \centering
\caption{
\jiazhao{Comparison results of LM-Nav and its variants on the Val-Unseen split of RxR dataset. Here, ``$*$'' indicates the vanilla version.}
}
\scalebox{1}{
\setlength{\tabcolsep}{2.5mm}{
\begin{tabular}{l|cc}
\hline
Method & OS↑ & SR↑ \\ \hline \hline
LM-Nav* (GPT 3.5, CLIP) & 24.0 & 7.38 \\
LM-Nav (GPT 4, EVA-CLIP) & 18.3 & 9.89 \\
LM-Nav (Vicuna-7B, EVA-CLIP) & 15.7 & 8.20 \\
NaVid (Vicuna-7B, EVA-CLIP) & \textbf{35.5} & \textbf{23.5} \\ \hline
\end{tabular}
    }
}
\label{tab:comp-lm-nav-1}
\end{table}

\textbf{Comparison with LM-Nav and its variant.} 
LM-Nav is a baseline which leverages discretized environment setting and off-the-shelf foundation models. Here, LM-Nav (GPT-3.5 and CLIP) employs GPT-3.5 for instruction decomposition and CLIP for landmark grounding. Additionally, we construct an ablation study, LM-Nav (Vicuna-7B and EVA-CLIP), substituting the foundation models of LM-Nav with those used in NaVid. We also implement a strong baseline, LM-Nav (GPT-4 and EVA-CLIP), which incorporates advanced foundation models as its building blocks. It is important to note that LM-Nav requires a pre-built landmark graph; hence, we utilized the ground truth landmarks from the Habitat simulator to construct this graph. Moreover, as the visual navigation model (ViNG) \cite{shah2021ving} employed by LM-Nav is not publicly accessible, we provide the ground truth shortest path to enable LM-Nav to navigate to each landmark effectively. We evaluate all methods on the RxR Val-Unseen split, where NaVid has not previously encountered the instructions and scenes during training.

The results are presented in Table~\ref{tab:comp-lm-nav-1}.
We observe that NaVid significantly outperforms LM-Nav on the RxR Val-Unseen split, despite that LM-Nav uses predefined oracle landmarks. This superior performance can be attributed that LM-Nav only focuses on landmarks, ignoring verbs and other directive commands (e.g., \textit{'turn right and walk to the chair'} or \textit{'turn around and face the sofa'}). Such a design neglects the spatial context of the instructions, potentially leading LM-Nav to incorrect landmarks with the same semantics as the intended targets. This problem is especially common in indoor environments where many objects may belong to the same semantic category. Moreover, models like CLIP and EVA-CLIP may struggle with image grounding in scenes with densely placed objects and obstructed viewpoints. We believe that NaVid, with its end-to-end training approach, demonstrates more adaptability and effectiveness for VLN tasks.

\begin{table*}[!t]\centering
\caption{
\jiazhao{Comparing in four diverse real-world environments scenes (\texttt{Meeting Room}, \texttt{Office}, \texttt{Lab}, and \texttt{Lounge}). Simple I.F. and Complex I.F. indicate the simple instruction following and complex instruction following tasks, respectively.}
}
\scalebox{1}{
\setlength{\tabcolsep}{1mm}{
\begin{tabular}{l|cccc|cccc|cccc|cccc}
\hline
 & \multicolumn{4}{c|}{\texttt{Meeting Room}} & \multicolumn{4}{c|}{\texttt{Office}} & \multicolumn{4}{c|}{\texttt{Lab}} & \multicolumn{4}{c}{\texttt{Lounge}} \\ \cline{2-17} 
 & \multicolumn{2}{c|}{Simple I.F.} & \multicolumn{2}{c|}{Complex I.F.} & \multicolumn{2}{c|}{Simple I.F.} & \multicolumn{2}{c|}{Complex I.F.} & \multicolumn{2}{c|}{Simple I.F.} & \multicolumn{2}{c|}{Complex I.F.} & \multicolumn{2}{c|}{Simple I.F.} & \multicolumn{2}{c}{Complex I.F.} \\ \cline{2-17} 
 & \multicolumn{1}{c|}{SR$\uparrow$} & \multicolumn{1}{c|}{NE$\downarrow$} & \multicolumn{1}{c|}{SR$\uparrow$} & NE & \multicolumn{1}{c|}{SR$\uparrow$} & \multicolumn{1}{c|}{NE$\downarrow$} & \multicolumn{1}{c|}{SR$\uparrow$} & NE$\downarrow$ & \multicolumn{1}{c|}{SR$\uparrow$} & \multicolumn{1}{c|}{NE$\downarrow$} & \multicolumn{1}{c|}{SR$\downarrow$} & NE$\downarrow$ & \multicolumn{1}{c|}{SR$\uparrow$} & \multicolumn{1}{c|}{NE$\downarrow$} & \multicolumn{1}{c|}{SR$\uparrow$} & NE$\downarrow$ \\ \hline \hline
Seq2Seq~\cite{krantz2020beyond}  & 4\% & 4.45 & 0\% & 7.21 & 0\% & 4.28 & 0\% & 6.92 & 0\% & 4.58 & 0\% & 6.61 & 0\% & 5.95 & 0\% & 6.82 \\
CMA~\cite{krantz2020beyond}  & 0\% & 4.27 & 0\% & 7.30 & 8\% & 4.62 & 0\% & 5.71 & 4\% & 4.35 & 0\% & 5.67 & 0\% & 4.63 & 0\% & 5.46 \\
WS-MGMap~\cite{chen2022weakly} & 52\% & 1.18 & 24\% & 2.20 & 60\% & 0.96 & 20\% & 2.94 & 44\% & 1.85 & 12\% & 3.18 & 48\% & 1.66 & 32\% & 2.88 \\
\textbf{Ours} & \textbf{92\%} & \textbf{0.55} & \textbf{56\%} & \textbf{0.98} & \textbf{84\%} & \textbf{0.63} & \textbf{48\%} & \textbf{0.71} & \textbf{76\%} & \textbf{0.83} & \textbf{40\%} & \textbf{1.89} & \textbf{88\%} & \textbf{0.72} & \textbf{44\%} & \textbf{1.37} \\ \hline
\end{tabular}
}
}
\vspace{-2mm}
\label{tab:real}
\end{table*} 


\begin{figure}[t]
\begin{center}
  \includegraphics[width=1 \linewidth]{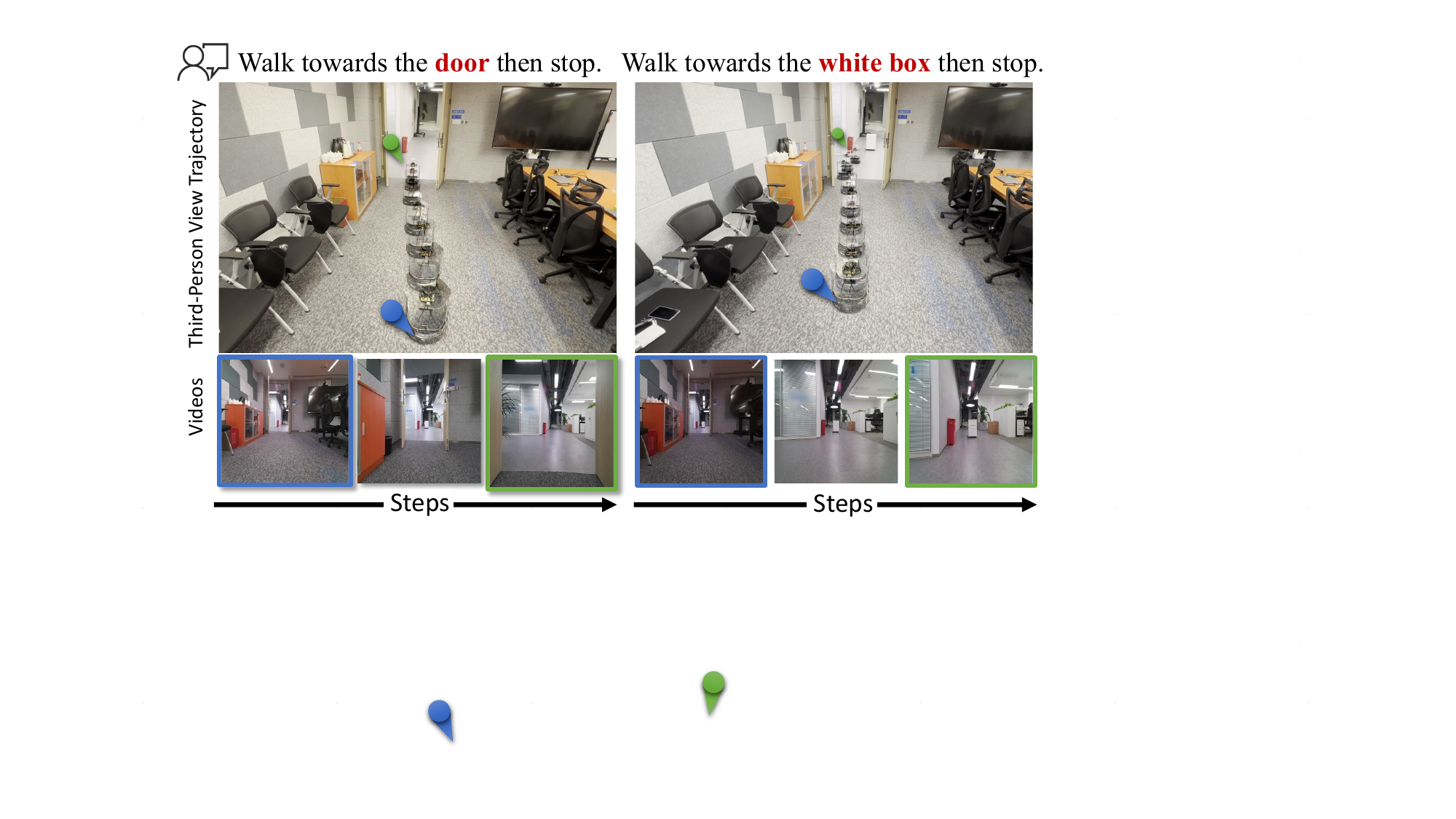}
\end{center}
   \caption{{Visual results of our method on \textbf{forward instructions} in real-world environments.} From top to bottom are instructions, third-person trajectories, and robot perspective videos.}
\label{fig:visual-com-forward}
\end{figure}

\subsection{Comparison on Real-world Environment.}

To further evaluate the generalizability of methods in more challenging situations, we conduct an extensive experiment in real-world environments. Here, We select two widely used baselines Seq2Seq and CMA, and a competitive method MS-MGMap (on Table~\ref{tab:comp-vlnce-r2r}). Each method is tested on four diverse environments with two types of instructions of different difficulties (25 simple instructions and 25 complex instructions for each scene). For a detailed description of real-world experiments please refer to Sec.~\ref{sec:exp-setup} and supplemental material. The results are presented in Table~\ref{tab:real}. Here, we find that our method showcases a significant improvement to all baseline methods. The end-to-end methods, Seq2Seq and CMA, suffer from extremely poor performance, which we believe is the sim-to-real gap between VLN-CE and the real world in terms of depth and color domain.
Compared to end-to-end methods, the map-based method, WS-MGMap, shows better performance by leveraging an ongoing semantics map. With careful parameters and algorithm adjustment, the map-based methods are widely regarded as a potentially robust strategy in the real world~\cite {zhang20233d,ramakrishnan2022poni}.
Nevertheless, our method, by exploring the generability of VLM in VLN, is able to complete most instructions (about $84\%$) and considerably complex instructions (about $48\%$), with only the requirement of RGB videos.

\begin{figure}[t]
\begin{center}
  \includegraphics[width=1 \linewidth]{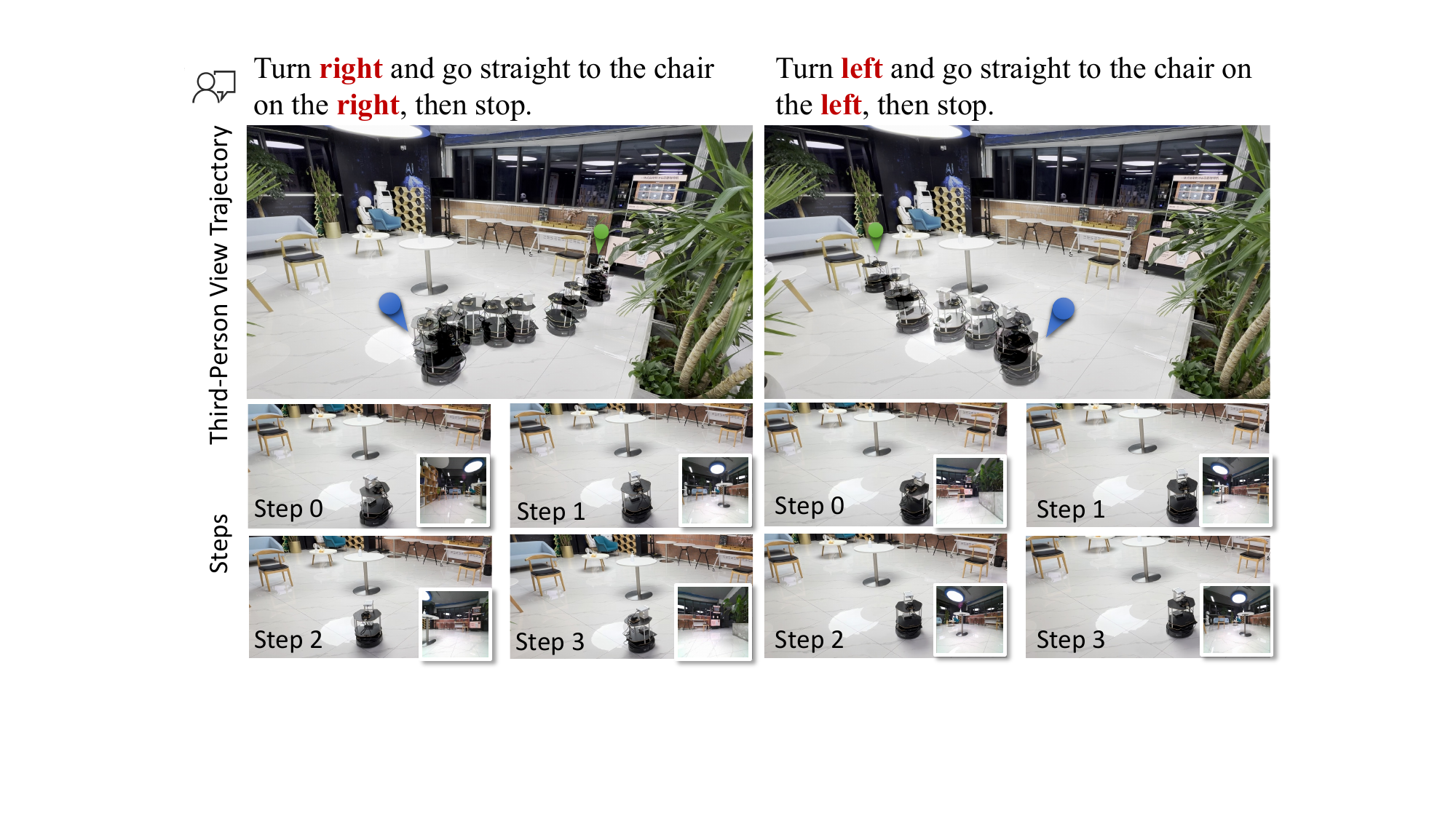}
\end{center}
   \caption{{Visual results of our method on \textbf{rotation instruction following} in real-world environments.} Our method can accurately park the robot near a specified target, despite initially being oriented towards a similarly classified object (a chair).}
\label{fig:visual-com-rotation}
\end{figure}

To further investigate the generalizability of \ours{} under different instructions, we showcase partial visual results of real-world experiments (Table~\ref{tab:real}), that require \ours{} to follow instructions with very specific landmarks and actions.

In Fig.~\ref{fig:visual-com-forward}, we give two similar instructions to drive the robot to move forward under \textit{different stop conditions}. We find that in both two cases, the robot can follow the instructions to move forward and successfully stop close to the given stop conditions. This demonstrates that \ours{} has the ability to understand the instructions and give correct actions for the robot to execute. Besides, we conduct a more challenging experiment by asking the robot to follow complex instructions. The visual results can be found in Fig.~\ref{fig:visual-com-rotation}, where the robot is required to turn orientation and move forward to stop near the given target. These instructions are particularly challenging because the robot is facing an object that shares the same semantic category as the target (chair). Nevertheless, our method is able to follow the instructions by turning the robot to the target orientation and stopping near the correct targets. The videos of these cases can be found in the attached supplemental material. 


We provide more visual results of our method in Fig.~\ref{fig:gallery} in real-world environments. More visual results can be found in the supplemental material. 

\begin{figure*}[t]
\begin{center}
  \includegraphics[width=1 \linewidth]{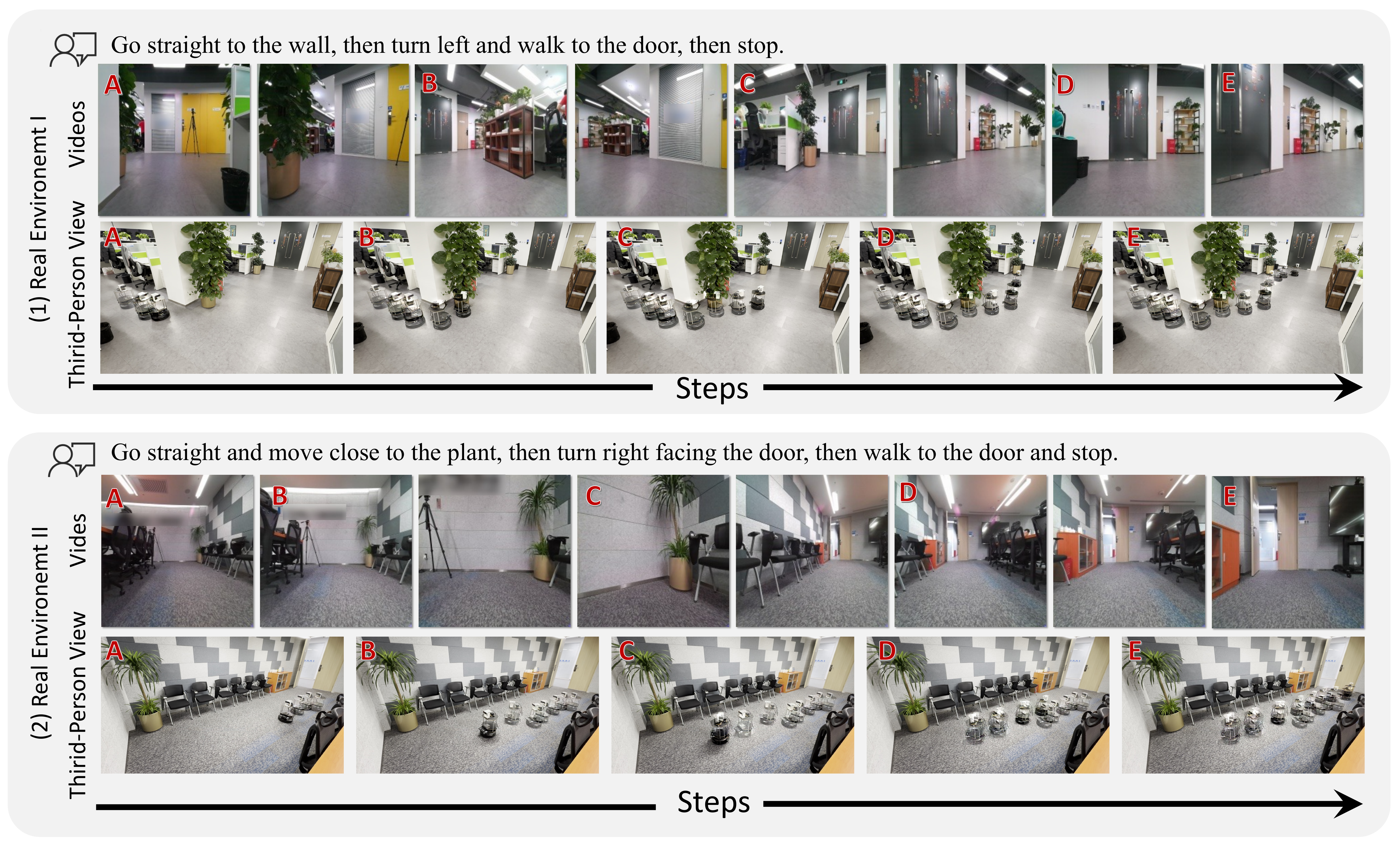}
\end{center}
   \caption{Visual results of our method in a real-world environment. In each real environment example, from top to bottom are instruction, robot perspective video, and third-person view video.}
\label{fig:gallery}
\vspace{-3mm}
\end{figure*}


\subsection{Ablation Studies.}
\begin{table}[!t]\centering
\caption{Ablation study on both training strategy, and architecture.}
\scalebox{0.9}{
\setlength{\tabcolsep}{0.95mm}{
\begin{tabular}{l|lccccc}
\hline
 & \multicolumn{6}{c}{VLN-CE R2R Val-Unseen} \\ \cline{2-7} 
 & \multicolumn{1}{c|}{Type} & TL & \textbf{NE}$\downarrow$ & \textbf{OS}$\uparrow$ & \textbf{SR}$\uparrow$ & \textbf{SPL}$\uparrow$ \\ \hline \hline
\multirow{3}{*}{\begin{tabular}[c]{@{}l@{}}Training\\ Strategy\end{tabular}} & \multicolumn{1}{l|}{No co-training} & 6.76 & 6.33 & 30.8 & 24.7 & 23.6 \\
 & \multicolumn{1}{l|}{No instruction reasoning sample} & 9.46 & 6.51 & 46.7 & 31.1 & 29.1 \\
 & \multicolumn{1}{l|}{No non-oracle navigation sample} & 8.73 & 5.82 & 46.4 & 34.2 & 32.0 \\ \hline
\multirow{4}{*}{Architecture} & \multicolumn{1}{l|}{No [NAV]} & 8.71 & 5.62 & 48.1 & 35.9 & 33.5 \\
 & \multicolumn{1}{l|}{No [HIS] and [OBS]} & 8.45 & 5.56 & 46.2 & 35.7 & 33.4 \\
 & \multicolumn{1}{l|}{Waypoints prediction} & 13.65 & 10.8 & 11.5 & 0.00 & 0.00 \\ \hline
\textbf{Ours} & \multicolumn{1}{l|}{Full pipeline} & 7.63 & \textbf{5.47} & \textbf{49.1} & \textbf{37.4} & \textbf{35.9} \\ \hline
\end{tabular}
}
}
\label{tab:ablation}
\end{table} 

To verify the effectiveness of each component in our method, we perform ablation studies on both training strategy and network architecture in Table~\ref{tab:ablation}. Through the experiment, we find that the co-tuning data is critical to the performance. This proved that lack of co-tuning data may cause the large foundation to lose generalizability. As expected, the instruction reasoning samples and Dagger nav-action samples show improvement. This inspires us that collecting more navigation-related data could further boost the performance of our method, which may be the follow-up direction of VLM for VLN. 
%
%
In the architecture ablation studies, we keep the content tokens (Instruction-queried token and instruction-agnostic token) and remove the special tokens: task identifier token \texttt{[NAV]} or observation identifier token \texttt{[HIS]} and \texttt{[OBS]}. We observe that the removal of special tokens leads to a noticeable performance drop, which proves the effectiveness of the special tokens.
The \textit{Waypoint prediction} variant replaces the $\cA_\text{action}$ with continuous location and orientation (see supplemental material for detailed implementation). 
However, the results demonstrate that direct output continuous location and orientation cause extreme challenges to VLMs, making VLMs difficult to learn navigation skills.

\begin{table}\centering
\caption{
\jiazhao{Ablation study on numbers of instruction-agnostic (visual) tokens per frame. We bold and underline the \textbf{best} and \underline{second best}, respectively.}
}
\scalebox{1}{
\setlength{\tabcolsep}{1.2mm}{
\begin{tabular}{l|cccccc}
\hline
 & \multicolumn{6}{c}{VLN-CE R2R Val-Unseen} \\ \cline{2-7} 
Tokens per frame & TL & NE↓ & OS↑ & SR↑ & SPL↑ & Avg. Time↓ \\ \hline \hline
(1) 1  tokens & 9.01 & 8.13 & 30.4 & 23.9 & 20.5 & \textbf{0.87s} \\
(2) 4  tokens & 7.63 & {\underline{ 5.47}} & {\underline{49.1}} & {\underline{37.4}} & {\underline{35.9}} & {\underline{1.22s}} \\
(3) 16 tokens & 9.10 & \textbf{5.38} & \textbf{51.1} & \textbf{38.0} & \textbf{36.1} & 2.72s \\ \hline
\end{tabular}
    }
}
\vspace{-2mm}
\label{tab:comp-visual-tokens}
\end{table} 

To evaluate the effectiveness of instruction-agnostic visual tokens comprehensively, we compare different settings of visual tokens. Specifically, we analyze NaVid with varying numbers of visual tokens per frame, noting that a higher token count retains more visual information. Here, we employ settings of 1, 4, and 16 of visual tokens, which are average pooled with strides of H, H/2, and H/4 of the feature map (H $\times$ H), respectively. The results are presented in Table~\ref{tab:comp-visual-tokens}.

From the results, we observe that utilizing more instruction-agnostic (visual) tokens per frame enhances performance, as the increased count of visual tokens provides richer visual information for action prediction. However, an increment in visual tokens also prolongs the inference time. We find that transitioning from setting (1) to setting (2) results in a $56.4\%$ increase in success rate and a $40.2\%$ increase in time costs. In contrast, moving from setting (2) to setting (3) yields only a $1.60\%$ increase in success rate alongside a substantial $122\%$ increase in time costs. This indicates that using four visual tokens per frame strikes an optimal balance, offering sufficient visual information for encoding navigation history while keeping reasonable computational costs.

\begin{figure}[t]
\begin{center}
  \includegraphics[width=1 \linewidth]{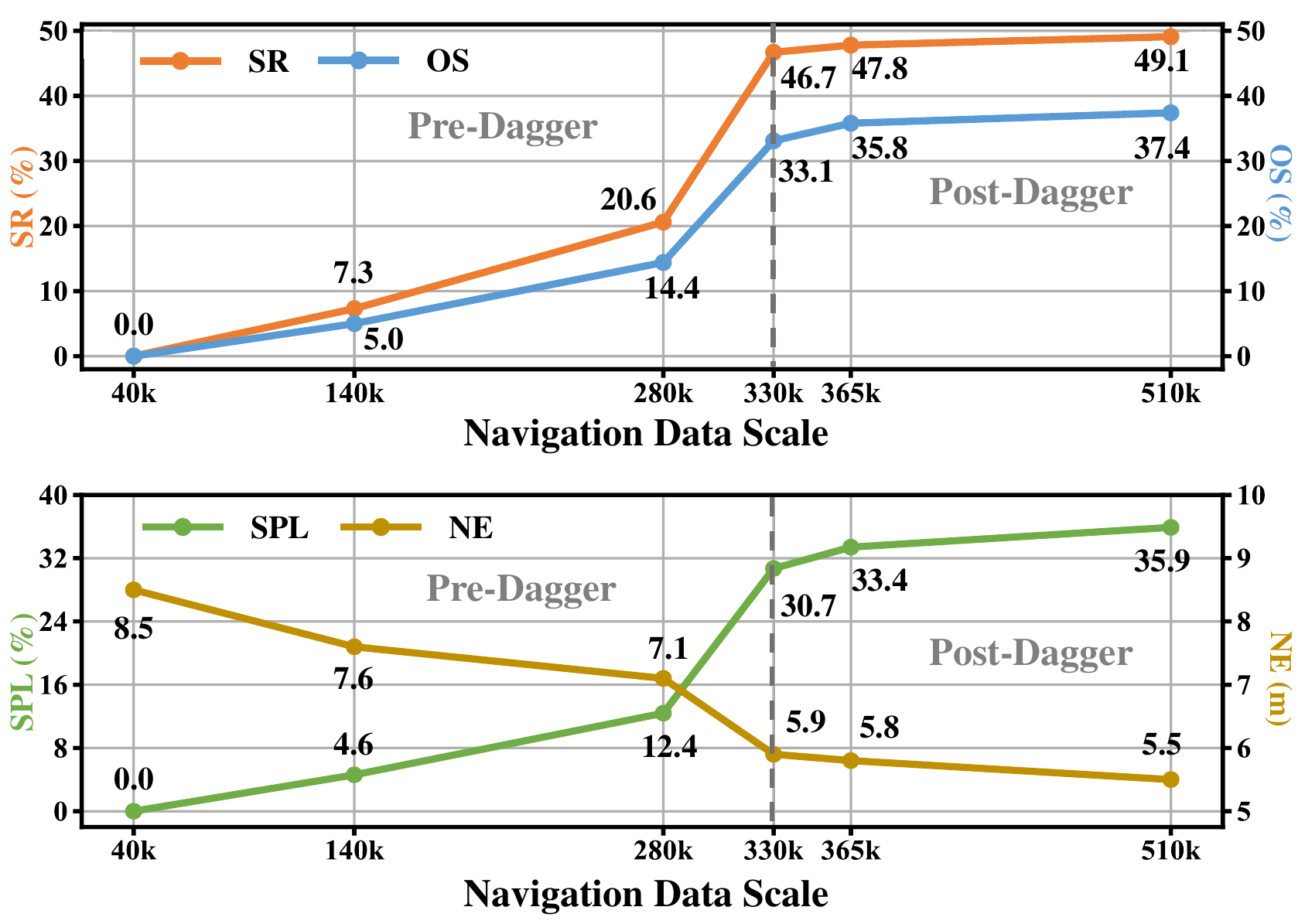}
\end{center}
   \caption{Performance analysis of our method on the VLN task across different data scales, showing the impact of data volume on SR, OS, SPL, and NE during pre-dagger and post-dagger training phases.}
\vspace{-4mm}
\label{fig:data-scale}
\end{figure}

We also conduct breakdown experiments to verify the importance of the navigation data in Fig.~\ref{fig:data-scale}. Here, we split the training into two phases, pre-dagger (0 to 330k) and post-dagger (330k to 510k). For pre-dagger, we randomly sample the navigation data for training. Here, we find that insufficient data (less than 280k) may lead to a slow improvement in terms of SR, OS, SPL, and NE. However, when the data increases to 330k, there is an obvious performance boost, indicating the model starts to master the VLN task. For the post-dagger phase, the improvements become minimal. The key reason is that the dagger on the R2R train split does not provide sufficient diverse environments or instruction information for the VLM to learn. 

\section{Discussion and Conclusion} 
\label{sec:conclusion}

In this paper, we propose a video-based VLM method, \ours{}, for the vision-and-language navigation task. NaVid achieves SOTA navigation performance without relying on odometers, depth sensors, or maps. Specifically, we extend a video-based VLM model to encode both historical and current navigation data by integrating self-defined special tokens. To learn the vision-instruction following ability, we collect 510k action planning samples from both R2R (320k) and Dagger (180k) and instruction reasoning samples (10k). The extensive experiments in simulator environments demonstrate that our method can achieve SOTA-level performance with only monocular videos as input. Besides, we deploy the \ours{} in real-world environments, showing generalizability to conduct VLN tasks in real worlds.

\textbf{Limitation.} Despite the promising results, \ours{} has several limitations. \textit{Firstly,} the computation cost of \ours{} causes a long latency problem, impacting navigation efficiency. A convincing way to mitigate this issue is to employ action chunk techniques~\cite{zhao2023learning} or quantization techniques~\cite{liu2023llava}. \textit{Secondly,} under very long horizon instructions, ~\ours{} may experience a performance drop due to the long context tokens problem~\cite{dong2023survey}. And the lack of high-quality long video annotation data intensified this problem. A viable approach to address these challenges is to incorporate more advanced large models ~\cite{li2023llama} as the backbone and utilize long-video data~\cite{wang2023internvid}.


\textbf{Future works.} We would like to further explore the potential of extending \ours{} to other embodied AI tasks, such as mobile manipulation~\cite{yenamandra2023homerobot, shridhar2020alfred, zhang2023gamma}. 
To achieve this, we would like to investigate various action designs that enable simultaneous control of the robot arm and mobile base. 
Additionally, it is critical to collect a dataset of annotated mobile manipulation videos to facilitate our model's understanding of both instructions and the physical interactions between objects and robots. 
Furthermore, we seek to enhance the efficiency of our model, allowing it to operate at higher speeds or on lower-cost hardware.




\section*{Acknowledgments}

We thank all the reviewers for their valuable comments and suggestions. We thank Qiying Yu for the fruitful discussions. We thank Lei Zhang, Feng Zhu, and Dongxu Yu for their assistance in building the real-world experimental system. This project is supported by BAAI and the joint laboratory of Peking University and Galbot.


\bibliographystyle{plainnat}
\bibliography{references}

\clearpage

\appendix

The supplemental material is organized as follows:

\begin{itemize}
    \item Sec.~\ref{sec:real-scenes} introduces implementation details.
    \item Sec.~\ref{sec:instructions} introduces more details about the instructions.
    \item Sec.~\ref{sec:robot} reports robot setup and implementation details.
    \item Sec.~\ref{sec:baselines} reports implementation details of baselines.
    \item Sec.~\ref{sec:quantitative} reports more experiment results.
\end{itemize}

\section{Real-world Experiments Setup}
\label{sec:real-setup}


We conduct real-world experiments following previous vision-and-language reports~\cite{habitat2020sim2real}, primarily focusing on indoor scenes, instructions, and robot setup.

\subsection{Indoor Scenes.}
\label{sec:real-scenes}

\begin{figure}[h]
\begin{center}
  \includegraphics[width=1 \linewidth]{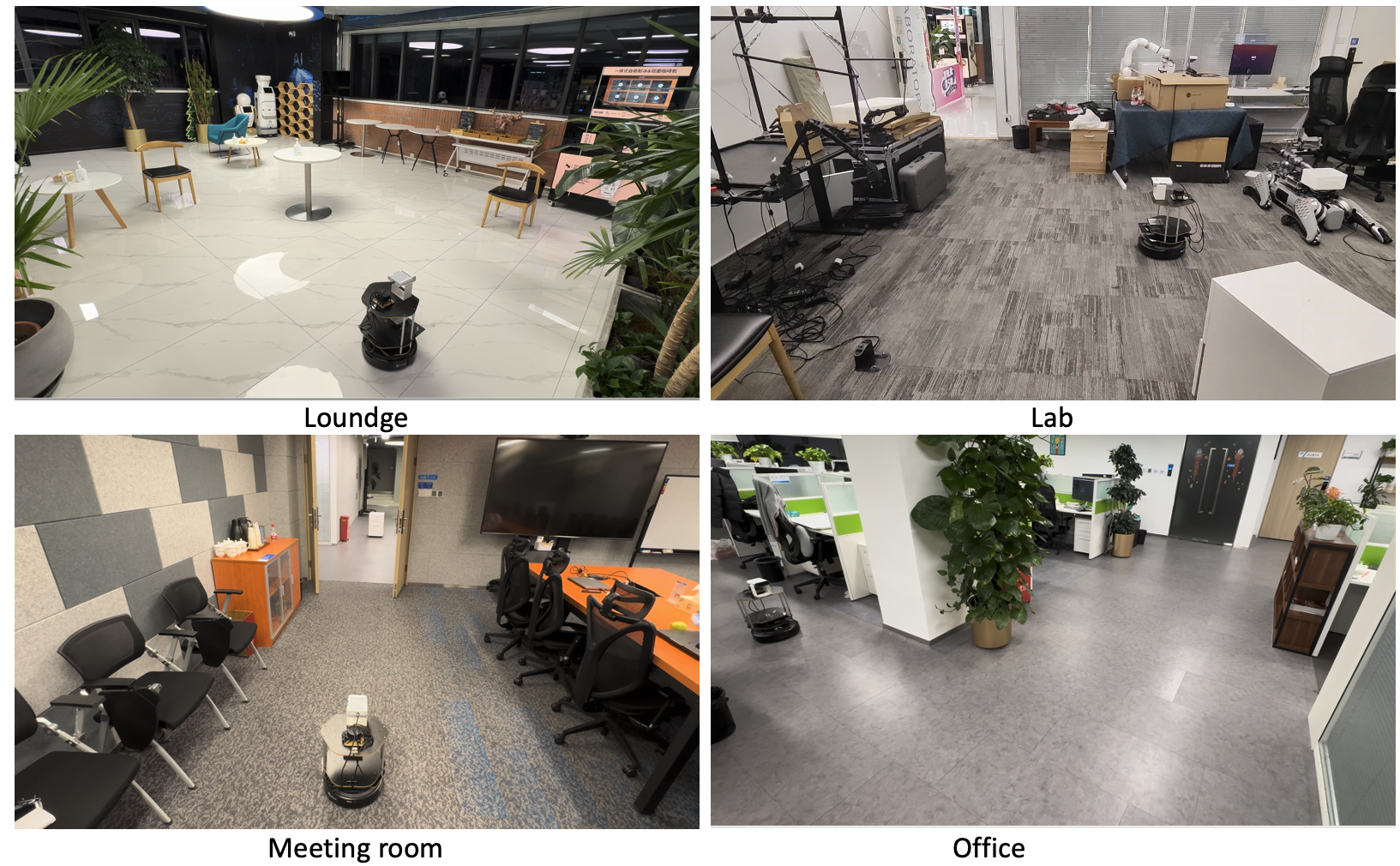}
\end{center}
   \caption{Indoor environments used in real-world experiments.}
\label{fig:scenes}
\end{figure}

We have chosen four challenging indoor environments, including \texttt{Meeting Room}, \texttt{Office}, \texttt{Lab}, and \texttt{Lounge}. These environments are crowded with objects of diverse styles and are subject to various lighting conditions. Visualizations of these environments are provided in Fig.~\ref{fig:scenes}.

\subsection{Instructions.}
\label{sec:instructions}

There are two types of instructions: (1) simple instructions, which require the agent to navigate to a single robot landmark and stop; (2) complex instructions, which require the agent to follow a series of simple instructions. All instructions are designed based on the objects present in the indoor scenes.

We list some examples of instructions used in the experiments. Here are some examples of simple instructions:
\begin{itemize}
    \item Walk to the plant.
    \item Move towards the white box then stop.
    \item Turn right and walk forward to the door then stop.

\end{itemize}

Here are some examples of complex instructions:
\begin{itemize}
    \item Go straight to the chair, then turn left/right to the stairs and stop.
    \item Move forward to the wall, then turn left. Walk through the
door, face the chair, and then stop.
\end{itemize}

\subsection{Robot setup.}
\label{sec:robot}

\begin{figure}[h]
\begin{center}
  \includegraphics[width=0.6 \linewidth]{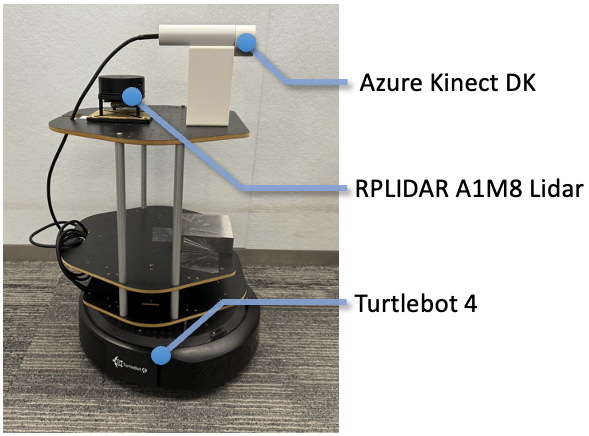}
\end{center}
   \caption{Real-world robot setup.}
\label{fig:robot}
\end{figure}

We provide a detailed description of our real robot (Fig~\ref{fig:robot}). Our robot is based on a Turtlebot 4, and we utilize the Azure Kinect DK to capture RGB and depth images (our method only utilizes RGB images). Additionally, we deploy a PRLIDAR A1M8 Lidar on top of the robot to capture a 1D laser point cloud, which is utilized by the lidar-odometry algorithm to compute location and orientation~\cite{macenski2020marathon2}.

\begin{figure}[h]
\begin{center}
  \includegraphics[width=1 \linewidth]{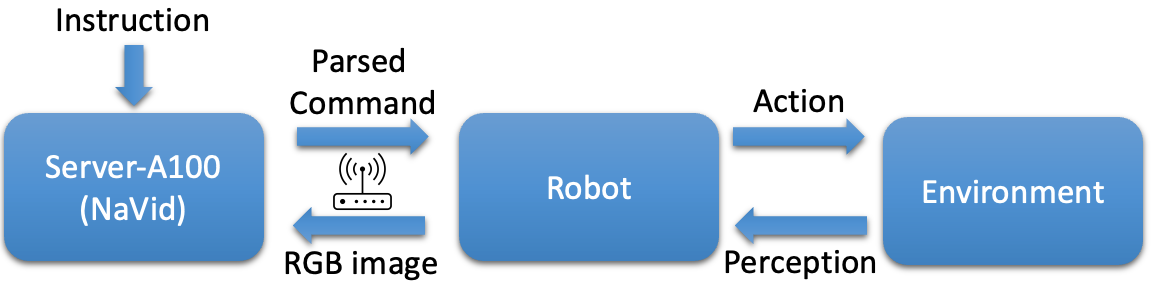}
\end{center}
   \caption{Pipeline of our navigation system.}
\label{fig:pipeline}
\end{figure}

Based on the above-mentioned robot, we design a pipeline for vision-and-language navigation with NaVid (Fig.~\ref{fig:pipeline}). NaVid is deployed on a server with an A100 GPU, which receives compressed images from the robot and sends back parsed commands through the Internet. The robot receives commands, such as "Turn left" or "Move forward", and drives the robot to execute the actions. During movement, the robot consistently tracks motion to confirm that rotation or forward movement is aligned with the commands.

Our pooling operation is performed with a specific stride on 2D feature maps. Specifically, we gridly split the 2D feature map from EVA-CLIP into grids and execute average pooling on each grid. For instance, we use a stride of H/2 to partition the H × H feature map into a 2 × 2 grid, where each grid leads to one token, culminating in a total of four tokens.

\begin{figure}[h]
\begin{center}
  \includegraphics[width=1 \linewidth]{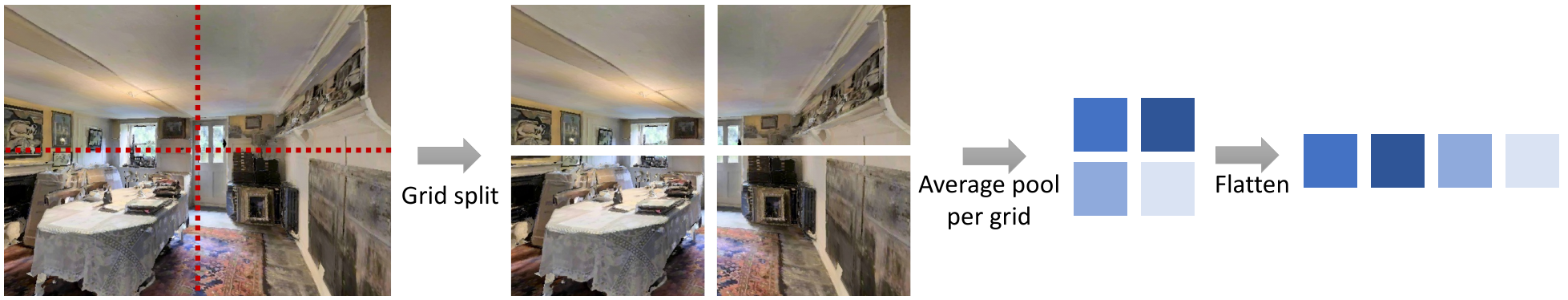}
\end{center}
   \caption{Illustration of our pooling operation.}
\label{fig:average-pool}
\end{figure}

\subsection{Baseline details}
\label{sec:baselines}

\textbf{LLaVA}\cite{liu2023llava} and \textbf{LLaMA-VID}\cite{li2023llama} are evaluated in Table III of the main paper. We directly use the models and weights from the openly released repositories\footnote{https://github.com/dvlab-research/LLaMA-VID}\footnote{https://github.com/haotian-liu/LLaVA}. However, we find that it is not able to consistently produce valid action outputs. To address this, we adopt prompt techniques following~\cite{zhou2023navgpt, long2023discuss}, incorporate CoT~\cite{wei2022chain} and utilize language-described historical information (captioned by LLaVA). Despite these efforts, most responses are related to scene descriptions and navigation skills, with valid navigation answers being rare.


\textbf{LLaVA-Nav} is a variant model trained using our navigation data. Since LLaVA can only process images and text, we follow existing work~\cite{zhou2023navgpt} and utilize text to describe the historical trajectory. Due to the token number limitation, we uniformly sample keyframes, including the first and last frames, and utilize LLaVA itself to describe these keyframes. We adopt the same output format as NaVid.


\textbf{Waypoint-NaVid} is a variant of NaVid that directly outputs a valid waypoint (location and orientation) for the next action. These waypoints are defined by the shortest path to follow the trajectory. For each prediction, we sample a valid waypoint within the field of view of the current observation. Despite using the same inputs as NaVid, waypoint prediction demonstrates extremely poor performance with a success rate of only $16\%$ (where success is defined as predictions with less than 30cm distance error and less than 30 degrees error), which leads to frequent failures.

\textbf{LM-Nav.} We utilize the ground truth landmarks from the Habitat simulator to construct this graph. Moreover, as the visual navigation model (ViNG) \cite{shah2021ving} employed by LM-Nav is not publicly accessible, we provide the ground truth shortest path to enable LM-Nav to navigate to each landmark effectively.

\subsection{Additional Experiments}
\label{sec:quantitative}


\textbf{Performance on object goal navigation}. To evaluate the performance of NaVid on tasks with sparser instructions, we conduct tests on the zero-shot object goal navigation task using the Habitat-Matterport 3D (HM3D) dataset~\cite{ramakrishnan2021hm3d}. Given the object category, this task challenges a robot to search for the target object within unseen environments. We directly modified the instruction to: '\textit{Search for {object}, move close to the {object}, and stop.}'. (Here, '{object}' is substituted with the category name of the target object). The criteria for success include: (1) achieving a Euclidean distance of no more than 1 meter from any instance of the target object category, and (2) ensuring that the object is visible from the stopping position. Importantly, our method has not been trained on the HM3D dataset and has not previously encountered the object goal search instruction format. We benchmark our results against mainstream open-set and zero-shot methods tailored for object goal navigation. The results are presented in Table~\ref{tab:comp-objectnav}.

\begin{table}\centering
\caption{
Comparison of object goal navigation task under open-vocabulary and zero-shot setting. The best and second-best results are highlighted in \textbf{bold} and \underline{underlined}, respectively.}

\scalebox{1}{
\setlength{\tabcolsep}{3mm}{
\begin{tabular}{l|cc}
\hline
Method & \multicolumn{1}{c}{SR↑} & SPL↑ \\ \hline \hline
WS-MGMap \cite{chen2022weakly} & 13.1 & 10.4 \\
ZSON \cite{majumdar2022zson} & 25.5 & 12.6 \\
GoW \cite{gadre2023cows} & 32.0 & 18.1 \\
ESC \cite{zhou2023esc} & \textbf{35.5} & \textbf{23.5} \\
NaVid & \multicolumn{1}{l}{{\underline{32.5}}} & \multicolumn{1}{l}{{\underline{ 21.6}}} \\ \hline
\end{tabular}
    }
}
\vspace{-3mm}
\label{tab:comp-objectnav}
\end{table} 

From the results, we observe that NaVid can outperform mainstream baselines such as Gow \cite{gadre2023cows} and ZSON \cite{majumdar2022zson}, despite not being specifically designed for object goal navigation, thus demonstrating robust performance in tasks with sparse instructions. In comparison to ESC \cite{zhou2023esc}, which utilizes ground truth location, orientation, and depth data, our method solely relies on RGB video for action prediction and yet achieves comparable performance (in terms of SPL). When compared to WS-MGMap \cite{chen2022weakly}, which is also trained on the VLN dataset, our method demonstrates significant improvements, underscoring the generalizability of our approach.

\textbf{Single-step prediction.} The sample for training NaVid is composed of a sequence of images and a step prediction of the next movements. The single-step prediction results can directly reflect the quality of robot learning. After training on all navigation samples (as described in the main paper), we evaluate the model on the new samples created from R2R Val-Uneen. We evaluate the success rate (correct prediction /number of samples), stop success rate (correct stop/number of stop samples), angle error (average angle degree error of correct rotation prediction) and distance error (average distance error of correct rotation prediction). Here, our method achieves $69.4\%$ success rate, $52.1\%$ stop success rate, $6.82$ (degree) angle error, and $16.7$ (centimeter) average distance error. We find this single-step performance leads to a $37.4\%$ success rate under the full vision-and-language task. The reason behind this is that the instruction following trajectory can be completed with different sequences of sing-step actions. Nevertheless, the model can not exactly follow the oracle trajectories, it can follow instructions with sub-optimal trajectories.

\begin{figure*}[h]
\begin{center}
  \includegraphics[width=1 \linewidth]{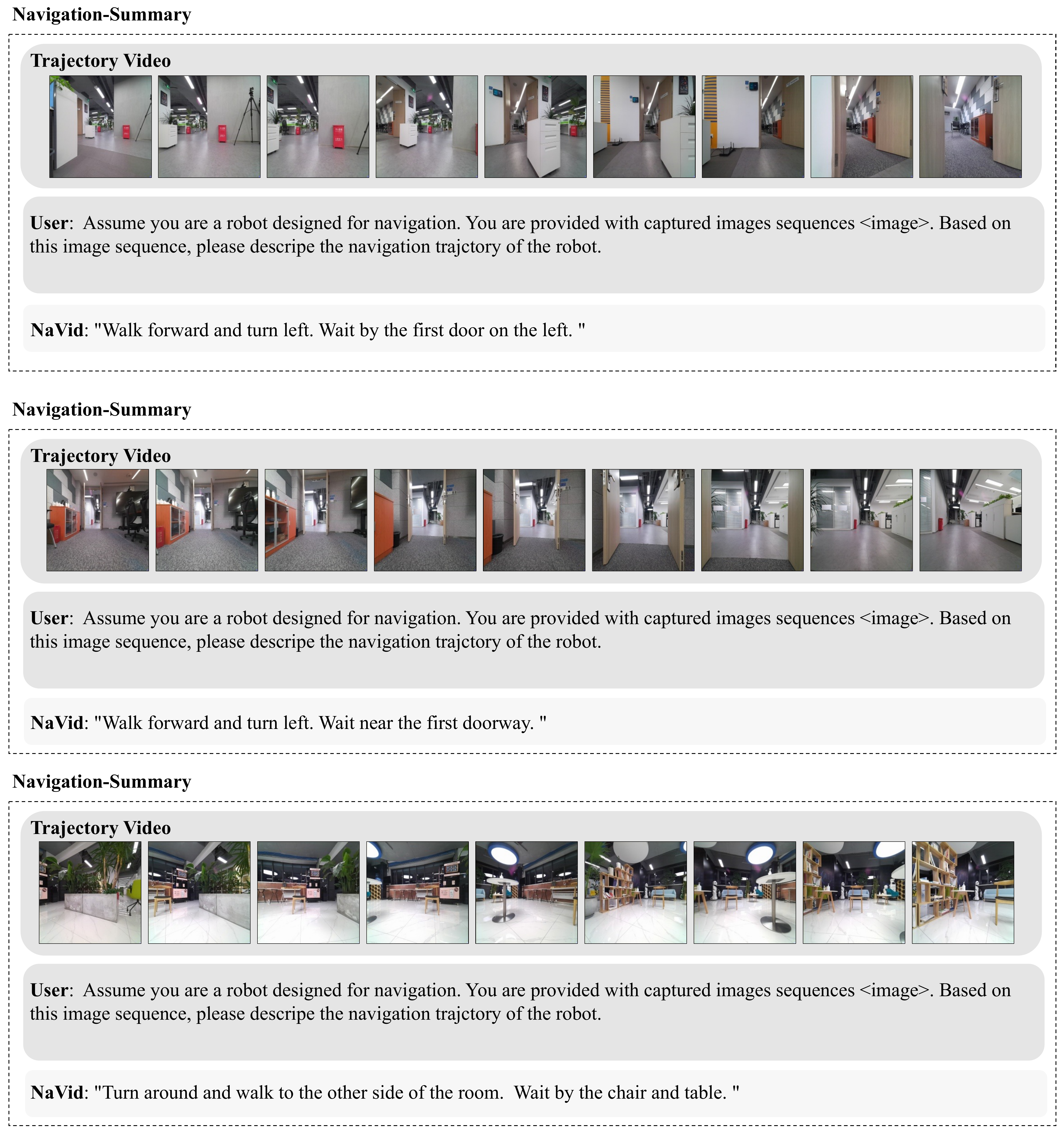}
\end{center}
   \caption{(1/2) Caption results visualization.}
\label{fig:caption-1}
\end{figure*}

\begin{figure*}[h]
\begin{center}
  \includegraphics[width=1 \linewidth]{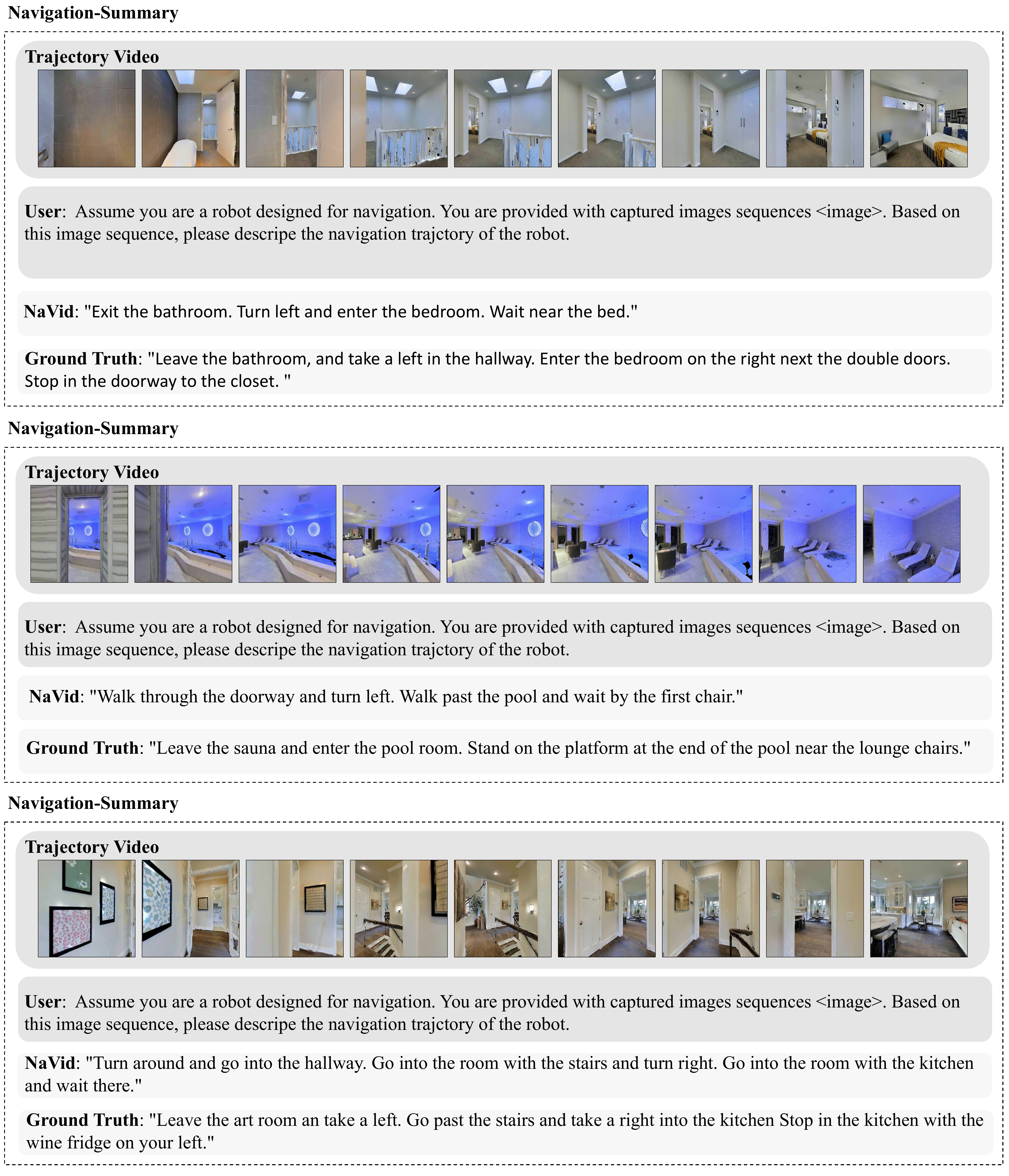}
\end{center}
   \caption{(2/2) Caption results visualization.}
\label{fig:caption-2}
\end{figure*}

\textbf{Trajectory summary.} As mentioned in the main paper (Sec. IV-C), the training samples include 10k instruction reasoning samples. Even the 10k samples are relatively low for a large language model to fully master the skill, but this indicates our model is able to do the trajectory summary task. Therefore, we test our model on diverse trajectories, including both real-world and simulator environments (shown in Fig.~\ref{fig:caption-1} and Fig.~\ref{fig:caption-2}). We observe that our model can briefly describe the trajectory, demonstrating that our model can understand its motions.


\section{Visual Experiments}
\label{sec:visual}

\begin{figure*}[h]
\begin{center}
  \includegraphics[width=1 \linewidth]{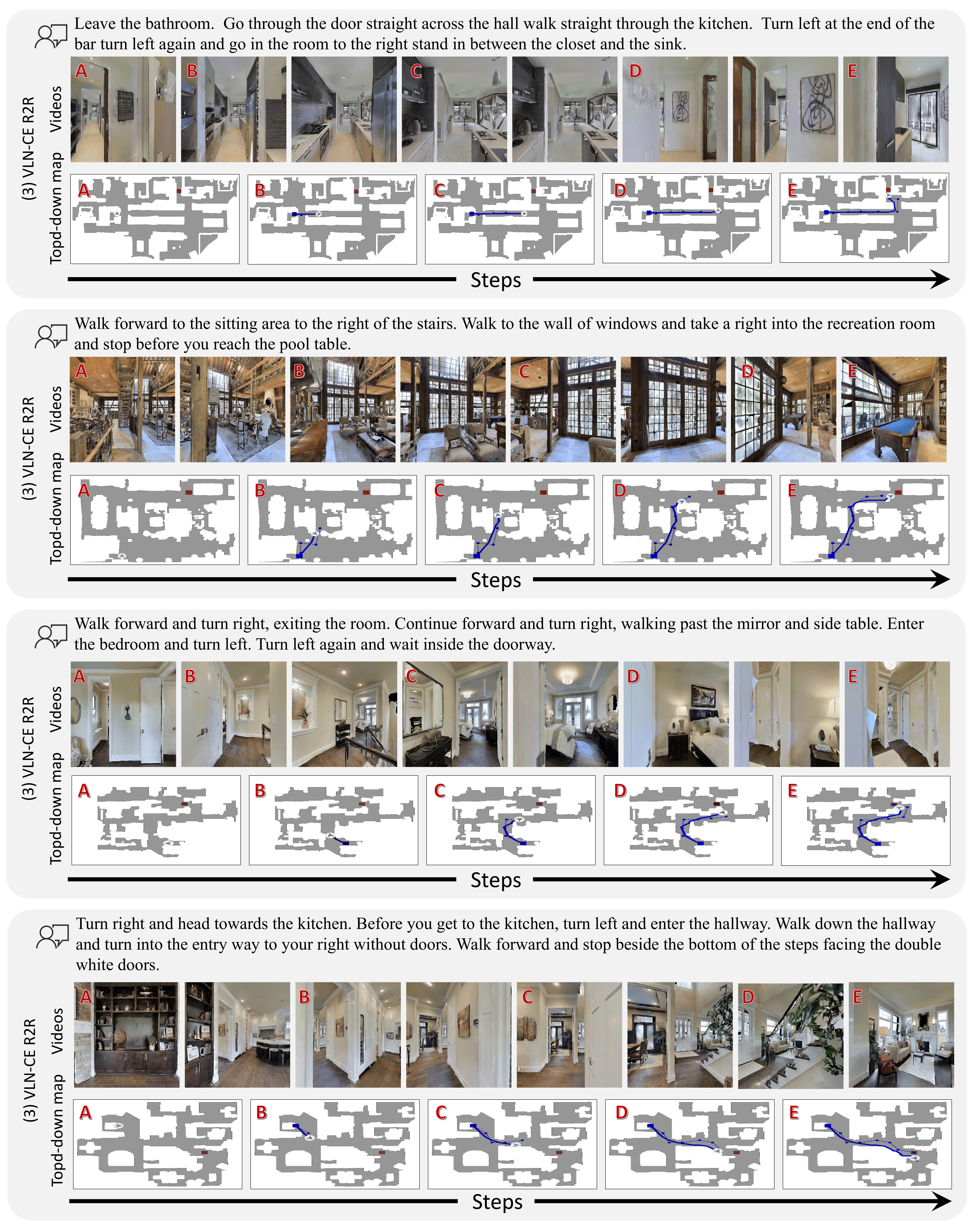}
\end{center}
   \caption{(1/2) R2R data visualization.}
\label{fig:r2r-1}
\end{figure*}

\begin{figure*}[h]
\begin{center}
  \includegraphics[width=1 \linewidth]{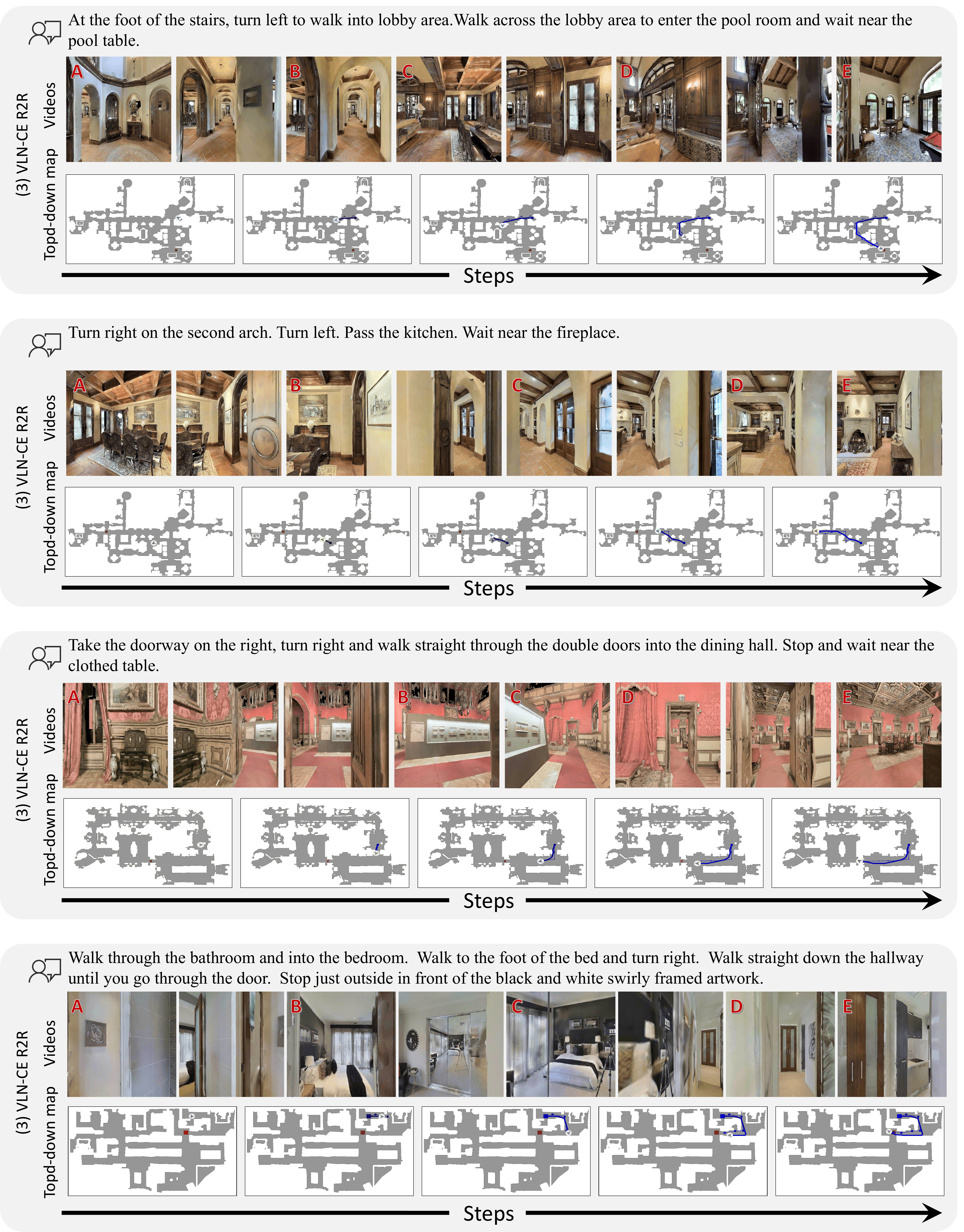}
\end{center}
   \caption{(2/2) R2R data visualization.}
\label{fig:r2r-2}
\end{figure*}

\begin{figure*}[h]
\begin{center}
  \includegraphics[width=1 \linewidth]{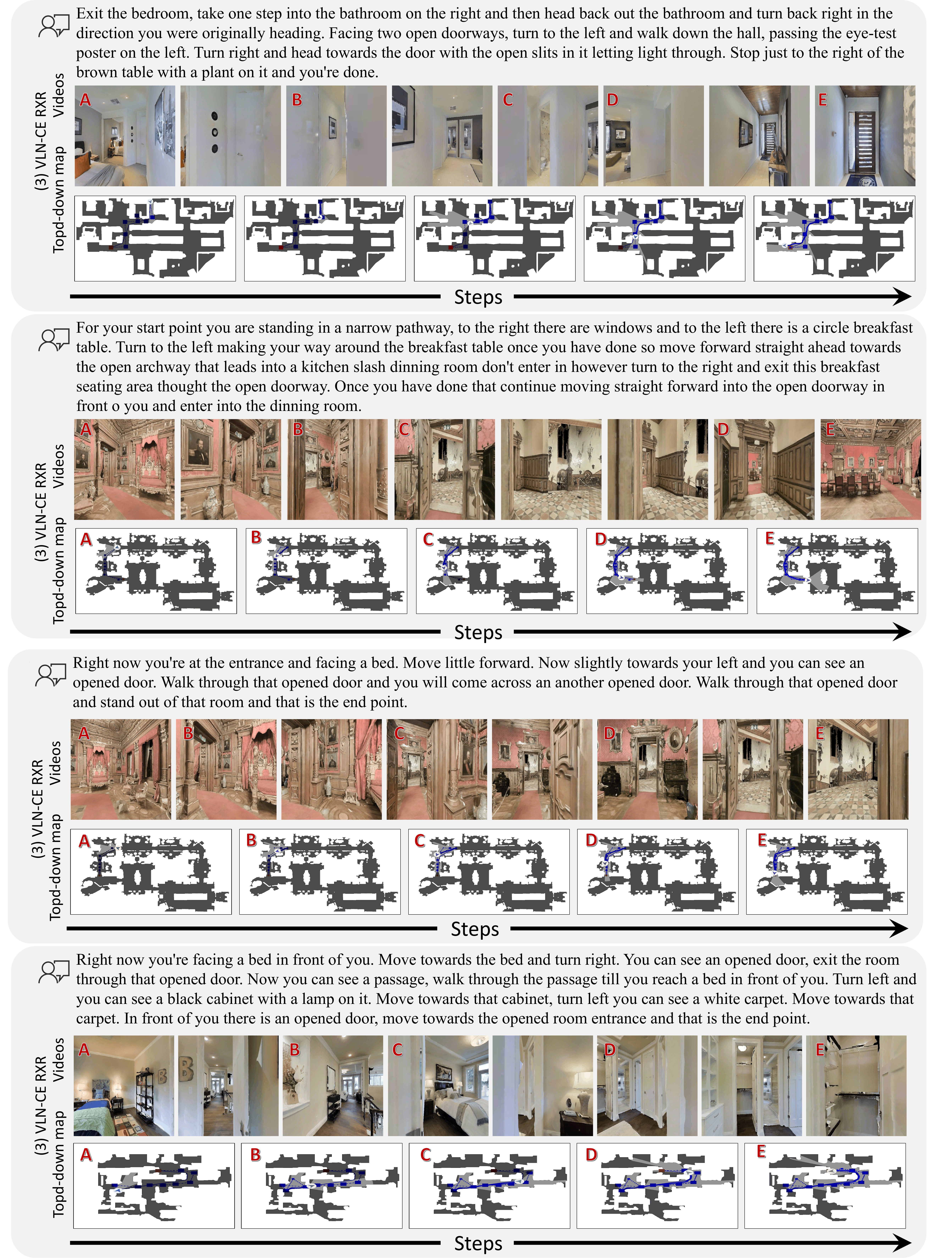}
\end{center}
   \caption{(1/2) RxR data visualization.}
\label{fig:rxr-1}
\end{figure*}

\begin{figure*}[h]
\begin{center}
  \includegraphics[width=1 \linewidth]{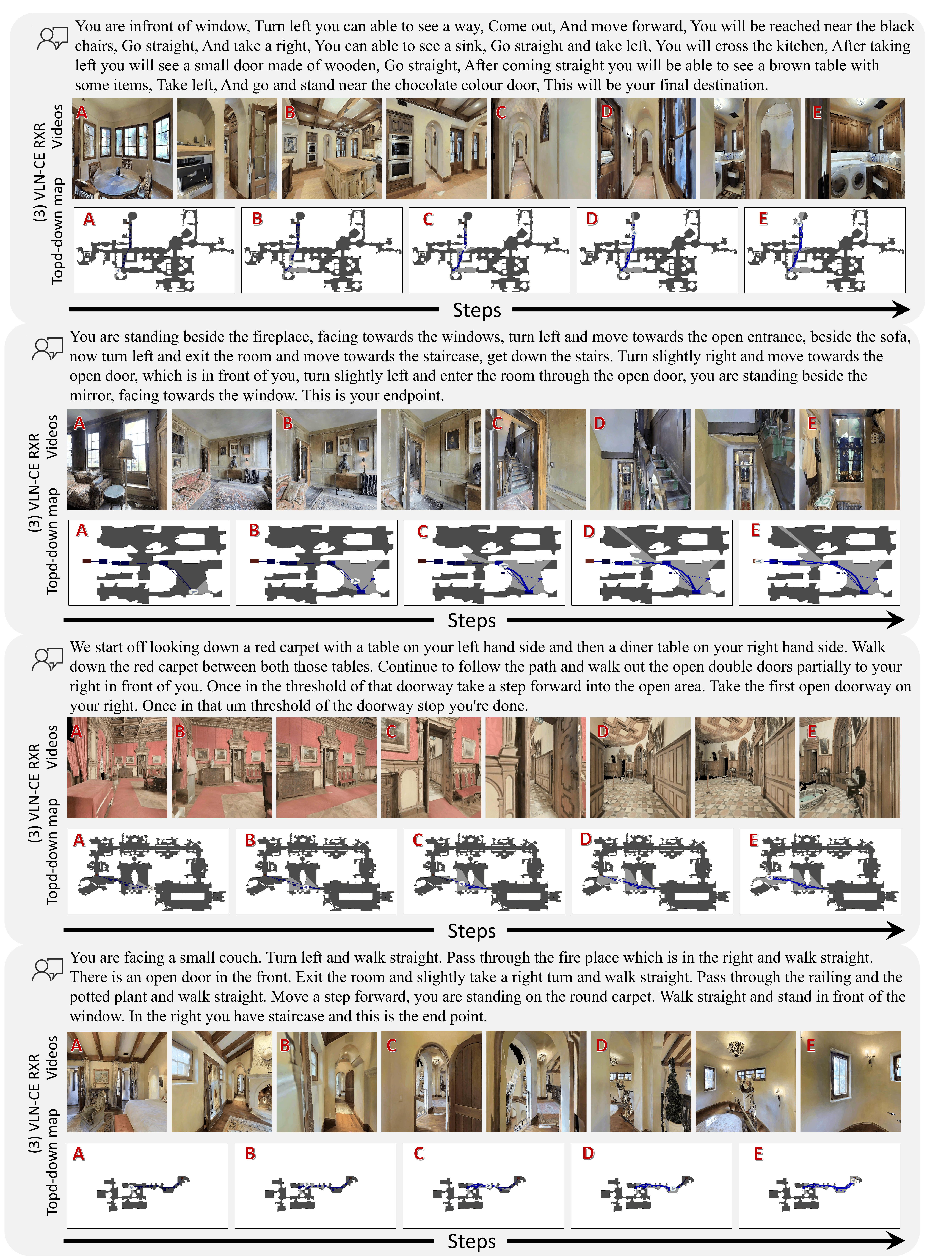}
\end{center}
   \caption{(2/2) RxR data visualization.}
\label{fig:rxr-2}
\end{figure*}

We propose extensive visual experiments of our method on simulator environments. Fig.~\ref{fig:r2r-1} and Fig.~\ref{fig:r2r-2} for VLN-CE R2R dataset and Fig.~\ref{fig:rxr-1} and Fig.~\ref{fig:rxr-2} for VLN-CE RxR dataset. For more visual results please refer to the attached video.

\end{document}